\documentclass[preprint,review,12pt]{elsarticle}
\usepackage{amssymb}
\usepackage{subcaption}
\usepackage{graphicx}
\usepackage{verbatim}
\usepackage{mathtools}
\usepackage{hyperref}
\usepackage{mathrsfs}
\usepackage{enumitem}
\usepackage{color}
\usepackage{booktabs}
\usepackage{natbib}
\usepackage{algorithm}
\usepackage{algpseudocode}
\usepackage[normalem]{ulem}
\usepackage{mathabx}

\usepackage{amsmath}
\usepackage{setspace}
\newcommand{\RNum}[1]{\uppercase\expandafter{\romannumeral #1\relax}}
\DeclareMathAlphabet{\mathpzc}{OT1}{pzc}{m}{it}

\def\pd{\partial}

\def\yel#1{\textcolor[rgb]{0.929,0.694,0.125}{#1}}
\def\ord#1{\textcolor[rgb]{0.635,0.078,0.184}{#1}}
\def\org#1{\textcolor[rgb]{0.850,0.325,0.098}{#1}}
\def\lgr#1{\textcolor[rgb]{0.733,0.837,0.594}{#1}}
\def\ogr#1{\textcolor[rgb]{0.466,0.674,0.188}{#1}}
\def\dgr#1{\textcolor[rgb]{0.233,0.337,0.094}{#1}}
\def\lbl#1{\textcolor[rgb]{0.500,0.7235,0.8705}{#1}}
\def\obl#1{\textcolor[rgb]{0.000,0.4470,0.7410}{#1}}
\def\dbl#1{\textcolor[rgb]{0.000,0.2235,0.3705}{#1}}
\def\linesolid{{--------}}

\journal{}

\begin{document}
	
	\begin{frontmatter}
		
		
		\title{Control of a fly-mimicking flyer in complex flow using deep reinforcement learning}
		
		\author{Seungpyo Hong \fnref{cont}}
		\author{Sejin Kim \fnref{cont}}
		\author{Donghyun You \corref{cor1}}
		\address{Department of Mechanical Engineering, Pohang University of Science and Technology, 77 Cheongam-Ro, Nam-Gu, Pohang, Gyeongbuk 37673, South Korea \vspace{-0.4in}}
		
		\fntext[cont]{Authors contributed equally.}
		\cortext[cor1]{Corresponding author.}
		
		\ead{dhyou@postech.ac.kr}
		 
		\begin{abstract}
		An integrated framework of computational fluid-structural dynamics (CFD-CSD) and deep reinforcement learning (deep-RL) is developed for control of a fly-scale flexible-winged flyer in complex flow.
        Dynamics of the flyer in complex flow is highly unsteady and nonlinear, which makes modeling the dynamics challenging.
        Thus, conventional control methodologies, where the dynamics is modeled, are insufficient for regulating such complicated dynamics.
        Therefore, in the present study, the integrated framework, in which the whole governing equations for fluid and structure are solved, is proposed to generate a control policy for the flyer.
        For the deep-RL to successfully learn the control policy, accurate and ample data of the dynamics are required.
        However, satisfying both the quality and quantity of the data on the intricate dynamics is extremely difficult since, in general, more accurate data are more costly.     
        In the present study, two strategies are proposed to deal with the dilemma.  
        To obtain accurate data, the CFD-CSD is adopted for precisely predicting the dynamics. 
        To gain ample data, a novel data reproduction method is devised, where the obtained data are replicated for various situations while conserving the dynamics.
        With those data, the framework learns the control policy in various flow conditions and the learned policy is shown to have remarkable performance in controlling the flyer in complex flow fields.
		\end{abstract}
		
		
	\end{frontmatter}
	
	\section{Introduction}
	From birds to insects, humans have been inspired by their marvelous flight performances and developing flapping-wing air vehicles (FWAVs) with such aerobatic feats~\cite{Phan2019}.
    Researchers initially started to control FWAVs in a quiescent fluid and have shown successful aviation in various flight modes~\cite{Karasek2018Delfly,Fei2019LearnExtreme,Phan2020Collision}.  
    Recently, for outdoor applications, trials on controlling FWAVs to hover in wind tunnels were made although the positioning accuracy was degraded with increasing wind velocity~\cite{Cunis2016precision,Karasek2019accurate,Chirara2017}.  
    On the other hand, increasingly smaller FWAVs have emerged~\cite{Ma2013} and the miniaturization trend is expected to continue even down to a fly scale~\cite{Pulskamp2012}. 
    Interest in precisely controlling small-scale FWAVs in complex flow is increasing.
    
    In order to elaborately control FWAVs, accurate prediction of the aerodynamic force exerted on vehicles should be preceded. 
    When calculating the aerodynamic force on a flapping wing, existing control methodologies mostly adopt quasi-steady models~\cite{Ellington1984,Sane2003}. 
    Quasi-steady models literally assume a steady motion of the wing and calculate the force on it. 
    This modeling approach has been widely used because it provides a closed-form solution enabling fast calculation.
    However, it has a few limitations on accurately calculating the aerodynamic force exerted on small-scale FWAVs aviating in complex flow.
    In fact, flow around a flapping wing is highly unsteady and nonlinear.
    In particular, it gets more complicated in complex flow, degrading the accuracy of the models~\cite{Bluman2017}.
    Furthermore, as FWAVs get smaller, the limitation becomes severe.
    It is well known that intrinsic instability is magnified for diminutive FWAVs~\cite{Ristroph2010,Ristroph2013,Phan2017,Teoh2012}
    and they are more sensitive to external flow due to their lighter weight, requiring more accurate prediction of the dynamics. 
    However, the high-frequency flapping and the rapid rotational motion of the wing observed in small-scale FWAVs are the representative cases the models become inaccurate~\cite{Altshuler2005,Bomphrey2017}.
    In addition, most of the wings have flexibility, but, it is often neglected in the models although it has a substantial influence on the aerodynamic force~\cite{Kang2011}.  
    Taking these aspects into account, accurately modeling the complicated dynamics occured in the listed situations is impractical.
    Therefore, in this study, we propose another control approach that does not model the dynamics.
    
    In recent years, as a representative data-driven controller which does not need a model for the dynamics, deep reinforcement learning (deep-RL) emerged with great prospects showing its superiority~\cite{Bucsoniu2018}. 
    For the deep-RL to successfully learn an optimal control policy, accurate and ample data of the dynamics should be provided.
    The accurate data can be obtained using computational fluid-structural dynamics (CFD-CSD) where the aerodynamic force on FWAVs and their motions including the deformation of the wing are calculated.
    The coupling framework of CFD and deep-RL was firstly used to control self-propelled swimmers~\cite{Gazzola2014} and, later, the framework successfully revealed the mechanism of harnessing vortices in fish schooling~\cite{Verma2018}.
    Deep-RL was also adopted for controlling a FWAV in a quiescent fluid, but a quasi-steady model was still used to identify the dynamics~\cite{Fei2019LearnExtreme}.
    The main reason why CFD-CSD is rarely combined with deep-RL is its huge computational cost, hindering deep-RL from getting enough data.
    In order to deal with the cost problem, we newly devise a data reproduction method, thus, enabling the coupling framework. 
    
    In the present study, an integrated framework of CFD-CSD and deep-RL is developed to construct an optimal control policy for a fly-mimicking flyer in complex flow.
    In addition, to resolve the cost problem of the CFD-CSD and gain ample data for the deep-RL, a novel data reproduction method is devised, where the obtained data from the CFD-CSD are replicated while conserving the dynamics. 
    In this study, the flyer is embodied by introducing a two-dimensional flexible flapping wing (\ref{sec:FSI}).
    Although the two-dimensional approximation has a few discrepancies with a three-dimensional flyer, since the major interactions of vortices exist in two dimensions~\cite{Birch2001span1,Shyy2007span2}, it is sufficient to see the superiority of the suggested framework.
    The developed framework learns the control policy by using multi-agent learning where several flyers aviate in different flow conditions and share their data with a single deep-RL.
    After the learning, the control policy is validated by controlling the flyer in complex flow with sudden disturbances.
    Lastly, the recovery behaviors of the flyer against the disturbances are analyzed. 
	
	\section{Methods} 
	\subsection{Coupling framework of CFD-CSD and deep-RL}
\begin{figure}[]
	\centering
	\includegraphics[width=\linewidth]{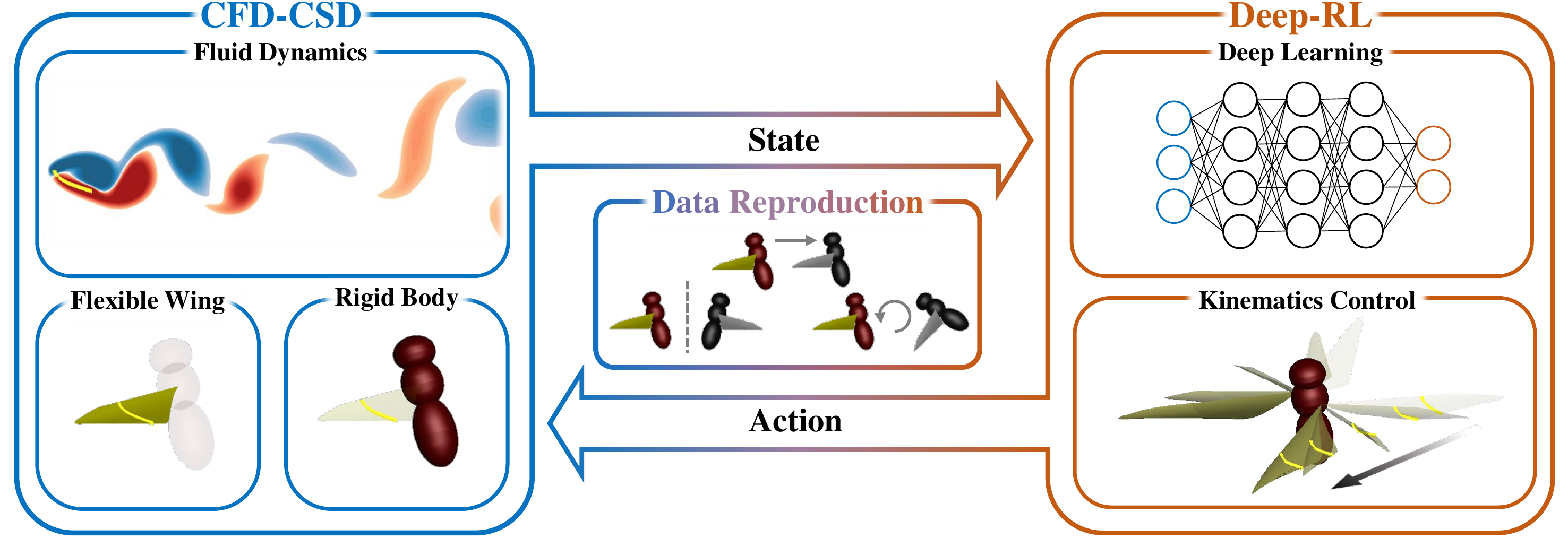}
	\caption{
    	Schematic of CFD-CSD and deep-RL coupling framework for generating a control policy of a fly-mimicking flyer.
	}
	\label{fig1:CFSD_DRL}
    \end{figure}

	Fig.~\ref{fig1:CFSD_DRL} shows the coupling framework of CFD-CSD and deep-RL.
    The aim of the framework is to generate a control policy for the wing kinematics of a fly-mimicking flyer.
    In order to learn the control policy, repetitive exchange of data (state and action) between CFD-CSD and deep-RL is conducted in every stroke.
    First, simultations of interaction among a fluid, a flexible wing, and a rigid body are conducted during a stroke.
    At the end of the stroke, a state such as the flyer's position and velocity is provided to the deep-RL.
    Then, the deep-RL determines an action which is the kinematics of the wing for the next stroke.
    During the CFD-CSD calculation for the next stroke, the deep-RL learns the control policy for a specific purpose set as a reward.
    Details of the CFD-CSD simulation and the deep-RL algorithm are provided in~\ref{sec:Methods}.         
    
    \subsection{Data reproduction method}
\begin{figure}[]	
	\centering
	\includegraphics[width=\linewidth]{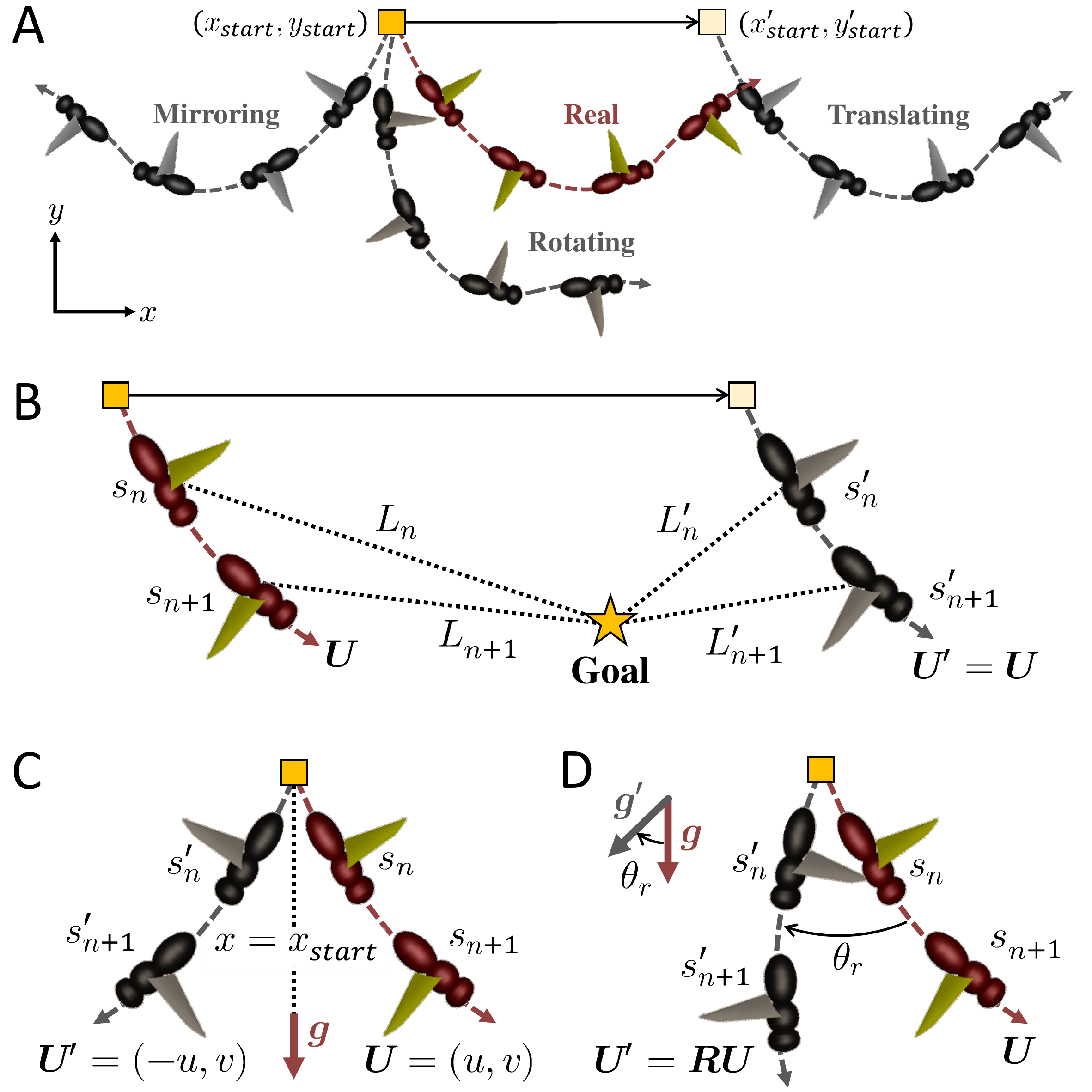}
	\caption{
    	Data reproduction methods. (A) Schematic of data reproduction methods. (B) Translation method. (C) Mirroring method. (D) Rotation method. $\bf{R}$ is a rotation matrix. Note that the direction of gravity rotates along with the flyer.
	}
	\label{fig:Repro}
    \end{figure}

    In order to overcome the data shortage problem and accelerate the learning procedure, a novel data reproduction method is developed.
    The main idea of the method is to augment real data experienced by the flyer to data different in terms of learning while conserving the dynamics of the flyer. 
    As shown in Fig.~\ref{fig:Repro}A, the reproduction methods are divided into three types: translation, mirroring, and rotation.
    
    Translation, as shown in Fig.~\ref{fig:Repro}B, relocates the starting point $(x_{start},y_{start})$ to different points $(x'_{start},y'_{start})$ so that the distance between the flyer and the goal changes.
    Mirroring, as shown in Fig.~\ref{fig:Repro}C, reflects data symmetrically with respect to the line $x=x_{start}$ so that the components like the flyer's velocity $\bf{U}$ are modified by shifting their sign in the $x$-direction.
    Since gravity is directed toward the negative $y$-direction, only mirroring in the $x$-direction is valid to conserve the dynamics.   
    The last method is rotation reproduction.
    However, due to gravity, merely rotating the flyer is not dynamically the same.
    In order to conserve the dynamics, gravity needs to rotate together as shown in Fig.~\ref{fig:Repro}D.
    In addition, gravity has to be included in the state of deep-RL because it is no longer constant. 
    
    Note that the present study employs an inertial frame while, in general, a body frame is prevalently adopted to represent the dynamics.
    This is because, by adopting the inertial frame, ample and diverse reproduced data can be obtained.
    In the case of the body frame, the mirroring and rotation reproductions merely alter the relative position of the flyer from the goal and all the other variables are unchanged; thus, only the number of data becomes larger without diversity. 
    In addition, the use of a body frame while controlling the flyer after learning does not matter because coordinate transformation can be easily done.
                               
    At this point, one might think that the real data and the reproduced data are the same in terms of learning since situations are dynamically identical.
    However, the deep-RL network does not recognize the dynamics itself, but, it only figures out the relationship among the numbers in the data.
    Thus, the reproduced data are considered as disparate data and, indeed, accelerate the learning process.
    The details of the data reproduction methods are described in~\ref{sec:DRM} and the advantage will be discussed in the following sections.
    
    \section{Results and discussion}  
    \subsection{Learning in a quiescent fluid}
    Note that all the upcoming variables are non-dimensionalized as described in~\ref{sec:NV}.
    As the first step, control of a flyer in a quiescent fluid is attempted.
    The flyer tries to learn how to stably aviate from the starting point $(-500,500)$ to the goal $(0,0)$ and hover.  
    If the flyer falls out of the computational domain $[-1000,1000]\times[-1000,1000]$, one episode is over.
    In a new episode, the flyer restarts from the starting point and the number of episodes given to the flyer is limited to three.
    Near the goal, a one-cent coin is overlapped to show hovering performance and to give a sense of the scale.
    The hovering-success condition is to keep the distance from the goal less than $5$ for more than $t=400$ (1 second). 
    If the condition is satisfied, the calculation is terminated.
    
\begin{figure}[]
	\centering
	\includegraphics[width=0.5\linewidth]{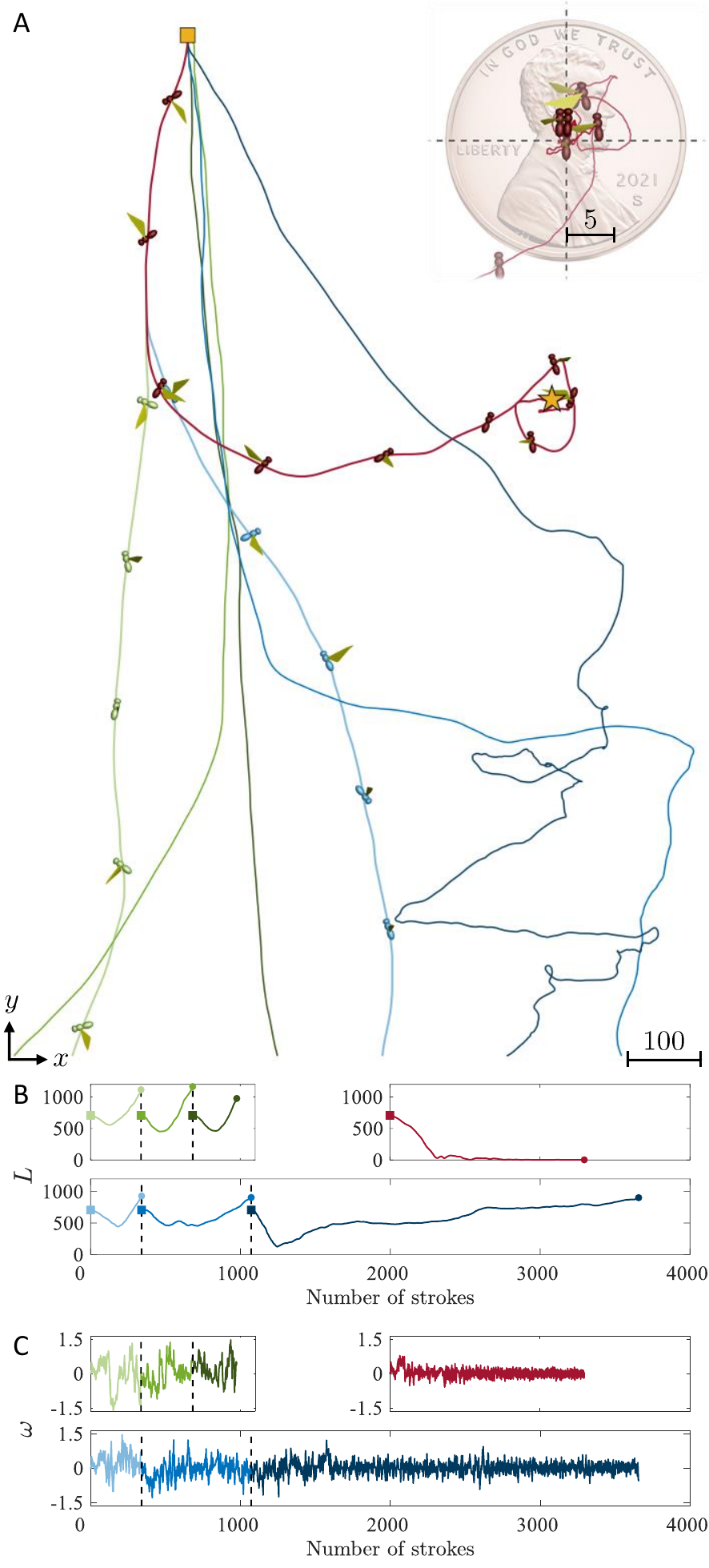}
	\caption{
    	Learning in a quiescent fluid. Without data reproduction: \lgr{\linesolid}, episode 1; \ogr{\linesolid}, episode 2; \dgr{\linesolid}, episode 3. With the translation and mirroring reproduction: \lbl{\linesolid}, episode 1; \obl{\linesolid}, episode 2; \dbl{\linesolid}, episode 3. With the translation, mirroring, and rotation reproduction: \ord{\linesolid}, episode 1. (A) Trajectories of flyers. (B) Distances between flyers and the goal in terms of the cumulative number of strokes (data). (C) Angular velocities of flyers. The objective of the flyers is to aviate from the yellow box to the yellow star and hover. For the episode 1 of all the cases, the flyers on the trajectories are depicted in $10$ times magnification with the time interval of $50$. Near the coin, the flyer is depicted in the real scale with the time interval of $100$. Note that the trajectory near the coin becomes clearer as the flight proceeds.
	} 
	\label{fig3:1vs1}
    \end{figure}

    Fig.~\ref{fig3:1vs1} shows the learning procedure with different combinations of data reproduction.
    As shown in the figure, the flyer which learns without the data reproduction falls down rapidly during all the episodes and cannot even stabilize its body.
    This clearly shows the difficulty of the present control where both stabilization and maneuver to the destination have to be handled simultaneously.
    The next case is the flyer learning with the translation and mirroring reproduction methods.
    It shows some improvements in that its angular velocity is gradually stabilized and the flyer withstands longer in the domain, but finally, it also fails.
    It seems that the data obtained from the three episodes are insufficient to learn the control policy.
    On the other hand, the flyer learning with all the reproduction methods succeeds in a single episode.
    It stabilizes its angular velocity quickly, aviates directly to the destination, and stably hovers.
    
    This result shows the apparent effect of the data reproduction method.
    In particular, the rotation reproduction is shown to have a significant role in accelerating the learning process.
    Note that, it can be realized by rotating gravity and including it in the state, even though gravity is fixed in the real world.
    
    \subsection{Multi-agent learning in uniform flow with various speeds}
    The aim of the present study is to make the controller that works in complex flow.
    In order to achieve it, the deep-RL should learn diversified flow composed of various directions and speeds. 
    The interesting thing is that learning flow with various directions can be omitted by virtue of the rotation reproduction. 
    To be specific, like gravity, external flow can be also rotated and reproduced in various directions.
    Therefore, although the flyer actually experiences unidirectional flow while learning, a myriad number of data with various flow directions are reproduced. 
    
    The remaining part to learn is various flow speeds.
    In the present study, multi-agent learning is utilized in which multiple flyers aviating in flow with different speeds share their experiences with a single deep-RL.
    In other words, the single deep-RL learns all the data obtained by the flyers and determines actions for them. 
    Specifically, four flyers are concurrently released in uniform flow with the different flow speeds in the positive $x$-direction ranging from $0$ to $1.76~(3m/s)$ with a constant interval.
    The range of the speed is selected to include dynamically harsh conditions with reference to the maximum flight speed of a fruit fly ($2m/s$)~\cite{Zhu2020Speed}.     
    
    While learning, one problem arises when it comes to sensing external flow.
    In the present study, the velocity of external flow is postulated to be sensed by an antenna~\cite{Fuller2014} on the head of the flyer and included in the state of the deep-RL. 
    However, during the early stage of learning, since the flyer is not capable of stabilizing its body and continuously rotates, the vortices shed from previous strokes distract the antenna from clearly detecting the external flow velocity. 
    Therefore, only during learning, the uniform flow velocity is used instead of the detected velocity for efficient and stable learning.
    After learning, with stabilized flight, the antenna-sensed velocity can represent the velocity of external flow.
    The unsuccessful result of the learning process using the antenna-sensed velocity is depicted in~\ref{sec:SRML1}.
    
\begin{figure}[]
	\centering
	\includegraphics[width=\columnwidth]{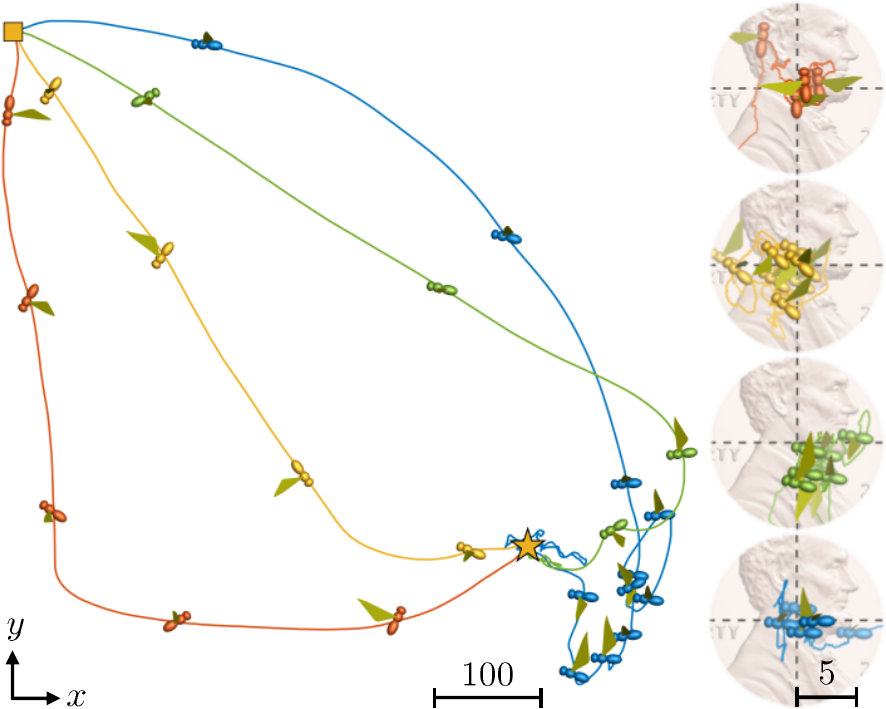}
	\caption{
    	Multi-agent learning in positive $x$-directional uniform flow with four different flow speeds: \org{\linesolid}, 0; \yel{\linesolid}, 0.59; \ogr{\linesolid}, 1.18; \obl{\linesolid}, 1.76. The flyers are depicted in 10 times magnification with the time interval of 50. Near the coins, the flyers are depicted in the real scale with the time interval of 100. Note that the trajectories near the coins become clearer as the flights proceed.
	} 
	\label{fig4:Multi}
    \end{figure}

    Fig.~\ref{fig4:Multi} shows the result of the multi-agent learning, in which four CFD-CSD calculations are performed for each flyer.
    If one of the flyers succeeds in hovering, the CFD-CSD calculation for that flyer is terminated, and the learning continues until all the flyers succeed.
    The hovering-success condition is the same as described in the previous section.
    As shown in the figure, all the flyers successfully reach the goal and hover.
    One can also see from the hovering flyers that, as the flow speed becomes faster, the flyers direct their heads more toward the upwind direction.
    
    Here, two queries can arise. 
    One is whether the resultant control policy from the learning can accurately control the flyer in external flow of untrained direction and magnitude.
    The other is whether the difference in using the boundary-imposed velocity during learning and the antenna-sensed velocity during control does not result in degraded performance of control.
    However, unlike these worries, the obtained control policy shows satisfying performance and the results can be seen in ~\ref{sec:SRML2}.
    
    \subsection{Control in complex flow with disturbances}
    In order to testify the performance of the learned control policy, a computational configuration for complex flow is set (\ref{sec:CC}).
    The control is performed without additional learning and, unlike the learning procedure, antenna-sensed velocity is used in the state.
    In complex flow, various missions are set.
    Firstly, the flyer is controlled to navigate from $(250,-250)$ to $(-250,-250)$.
    Then, when the distance from the goal is less than $5$, the position of the goal is suddenly changed from $(-250,-250)$ to $(-250,250)$.
    Again, when the flyer reaches the goal, the flyer is directed to move toward $(250,250)$.
    But, when the flyer goes through the line $x=125$, the goal is changed back to $(-250,250)$, which is intended to see the capability of changing its direction dramatically to the opposite. 
    In the same way, the goal is changed to $(0,0)$ when the flyer passes through the line $x=0$ to see the performance of a sudden $90^{\circ}$ turn. 
    After all, the flyer is controlled to hover at $(0,0)$.
    
    In addition, abrupt disturbances are added to show robustness of the control.
    At the middle of the aviation from $(250,-250)$ to $(-250,-250)$ and from $(-250,-250)$ to $(-250,250)$, intensive counter-clockwise body moment and negative $y$-directional body force are applied to the flyer, respectively.
    The disturbances are intended to represent the situations in which the flyer excessively loses its stability or it collides with something like a raindrop.
    In the case of the moment disturbance, $62.5$ times larger moment is applied than the moment used for the stability test of a fruit fly in Ristroph \textit{et al.}~\cite{Ristroph2010} during the same duration of $t=2$.
    In the case of the force disturbance, the force $300$ times larger than the flyer's weight is exerted on the flyer during $t=0.4$, referring to the experiment about the impact of a raindrop on a mosquito~\cite{Dickerson2012}.
    
    Fig.~\ref{fig5:PulseFly} shows the result of the controlled flight in complex flow with disturbances.
    As seen in the figure, the flyer shows off stable and agile maneuver in complex flow, quickly adapting to the confronted situations.
    Moreover, it also rapidly recovers its own pace after disturbances, showing robustness of the control.
    To sum up, the learned policy enables the flyer to aviate as intended in complex flow.
    
    \subsection{Recovery behaviors of the flyer against disturbances}
\begin{figure}[]
    \centering
	\includegraphics[width=1.0\textwidth]{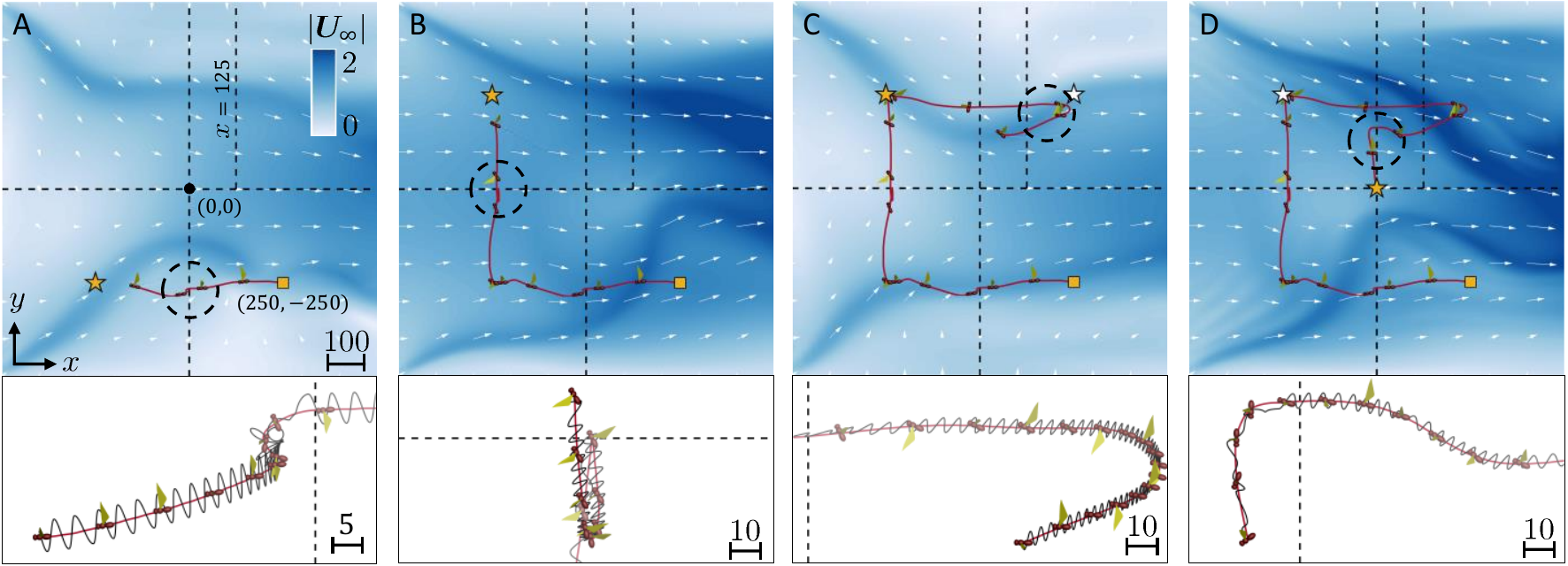}
	\caption{	
		Controlled flight of the flyer in complex flow with disturbances. The trajectory of the flyer during the flight: (A) from (250,-250) to (-250,-250) with the counter-clockwise body moment exerted at the line $x=0$; (B) from (-250,-250) to (-250,250) with the negative $y$-directional body force exerted at the line $y=0$; (C) from (-250,250) to (250,250) with a sudden change of the goal back to (-250,250) when the flyer passes the line $x=125$; (D) while flying to (-250,250) with a sudden change of the goal to (0,0) when the flyer passes the line $x=0$. The upper four figures show the entire computational domain and the background flow fields colored with the flow speed correspond to the moments at the end of the trajectories. The yellow stars are the goals at those moments and the white stars in (C) and (D) are the previous goals before the goal changes. The lower four figures are the enlarged parts marked as the black dashed circles in the upper figures. In the upper figures, the flyer is depicted in $10$ times magnification with the time interval of $50$. In the lower figures, the flyer is depicted in the real scale (A) or $2$ times magnification (B,C,D) with the time interval of $5$ (A,B,C,D). The black lines in the lower figures represent the traces of the wing's leading edge. Note that the trajectories in the lower figures become clearer as the flight proceeds.
	}
	\label{fig5:PulseFly}
    \end{figure}

    In Fig.~\ref{fig5:PulseFly}, four notable situations are briefly observed. 
    Among them, the first two situations including recovery behaviors against disturbances are analyzed in detail.
    
\begin{figure}[]
	\centering
	\includegraphics[width=0.6\linewidth]{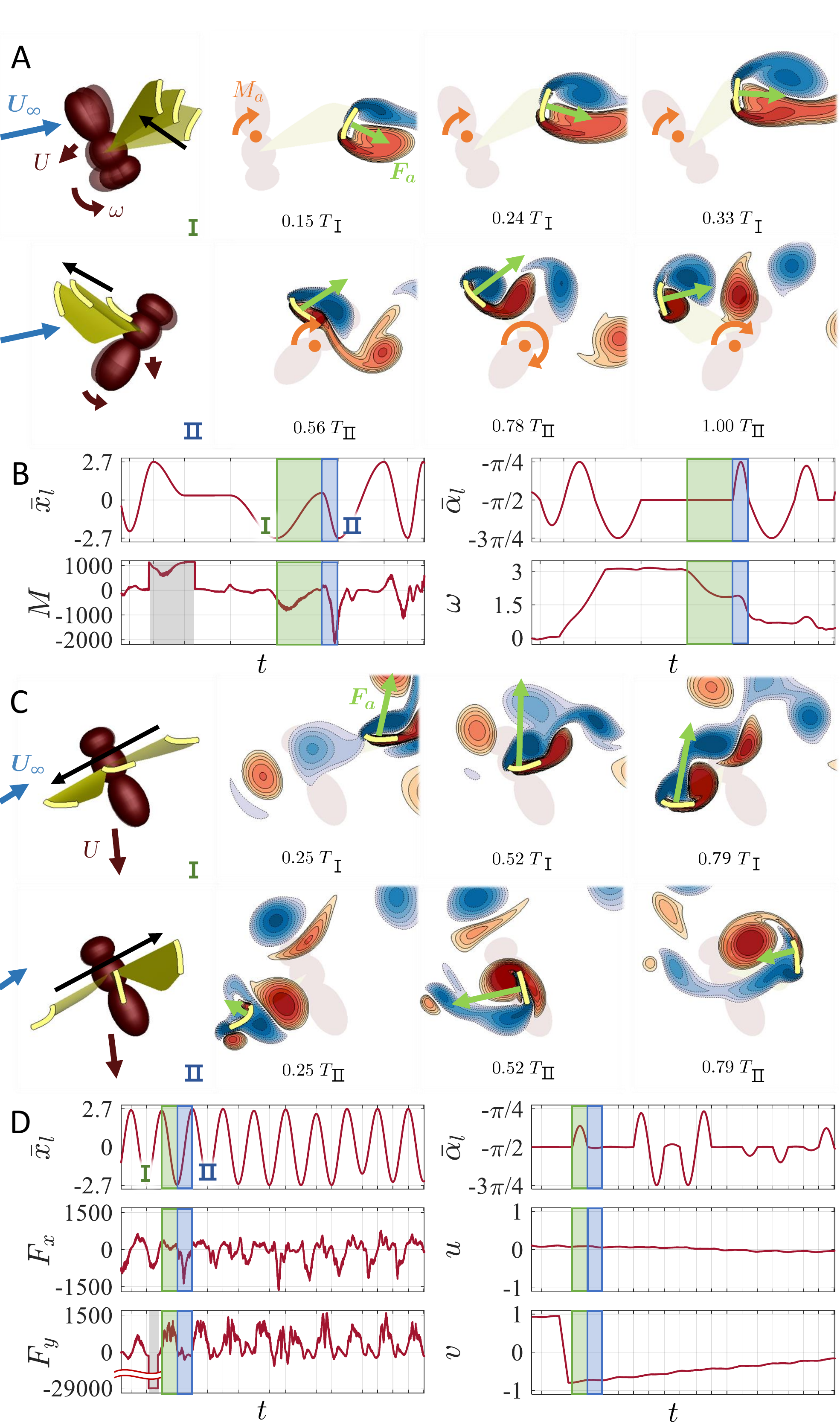}
	\caption{
		Analysis of recovery behaviors of the flyer against disturbances.
		(A) Motions of the flyer and ambient flow fields corresponding to the part~\RNum{1} and~\RNum{2} in (B), respectively. (B) Time histories of the wing kinematics, total moment exerted on the flyer, and angular velocity of the flyer. (C) Motions of the flyer and ambient flow fields corresponding to the part~\RNum{1} and~\RNum{2} in (D), respectively. (D) Time histories of the wing kinematics, total $x$- and $y$-directional forces exerted on the flyer, and $x$- and $y$-directional velocities of the flyer.
		The flow fields in (A) and (C) are depicted using vorticity where red and blue color refer to counter-clockwise and clockwise vortex, respectively.
		All the vectors are expressed in proportion to their magnitudes.
		$T_{\text{\RNum{1}}}$ and $T_{\text{\RNum{2}}}$ refer to the period of time to take for the stroke~\RNum{1} and~\RNum{2}.
		The gray-colored parts in (B) and (D) refer to the duration of exerting the disturbances on the flyer: positive (counter-clockwise) body moment (B) and negative $y$-directional body force (D). $\bar{x}_l$ and $\bar{\alpha}_l$ represent the position and the angle of the wing at the leading edge.
	} 
	\label{fig6:Torque}
    \end{figure}

    The first situation is to stabilize after the counter-clockwise body moment. 
    In the stabilization process, two behaviors are observed.
    The first one is to utilize the drag on the wing arising from the high relative velocity composed of high-velocity external flow and rapid rotation of the body.		
    As shown in the first row of Fig.~\ref{fig6:Torque}A and Fig.~\ref{fig6:Torque}B, the flyer erects its wing vertically to the external flow direction while slowly stroking up so that the clockwise moment consecutively exerts on the flyer. 
    The second one is to use the aggressive rotational motion of the wing.
    As can be seen from the second row of Fig.~\ref{fig6:Torque}A and Fig.~\ref{fig6:Torque}B, the flyer maximizes the rotational motion and frequency of the wing for the downstroke.
    In addition to the active kinematics, due to the passive deformation of the wing, the wing is further aligned in the direction where strong moments can be generated.
    As a result, strong clockwise moments arise during the stroke, rapidly stabilizing the angular velocity of the flyer.
    In the two behaviors mentioned above, the flyer is shown to utilize high-velocity external flow for its stabilization as opposed to the common sense that external flow might distract the stabilization.
    Moreover, the flexibility of the wing is also shown to have a significant role in stabilization.
    Surprisingly, within only two strokes, the flyer could almost stabilize its angular velocity.
    
    The second situation is to recover the velocity of the flyer after the negative $y$-directional body force.
    Time histories of the forces in the $x$- and $y$-directions in Fig.~\ref{fig6:Torque}D shows that the downstroke and the upstroke take different roles in the recovery process.
    It shows that the downstroke (part~\RNum{1}) and the upstroke (part~\RNum{2}) alternately generate forces in the $y$- and $x$-directions, respectively.
    During the downstroke, as shown in the first row of Fig.~\ref{fig6:Torque}C, the flyer lays its wing horizontally to generate the positive $y$-directional force, utilizing the high relative velocity in the positive $y$-direction.
    On the other hand, as shown in the second row of Fig.~\ref{fig6:Torque}C, the flyer stands the wing vertically to generate the negative $x$-directional force, compensating the positive $x$-directional external flow.
    Thus, the flyer can deal with both gravity and external flow well by dividing the role of each stroke, showing the systematic control of the learned policy.
    
    Throughout this section, an analysis is made on how the flyer behaves in the recovery process with existence of external flow.
    In addition, in order to rule out the effect of external flow, the same mission described in the previous section is performed in a quiescent fluid and the discrepancies are discussed in~\ref{sec:CB}.
    Enabling the analysis of the flyer's behavior considering the ambient flow field is an exclusive advantage of the proposed CFD-based framework, which makes deep understanding on the control mechanisms viable.
    
    \section{Concluding remarks}
    In the present study, an integrated framework of CFD-CSD and deep-RL is developed to extract a control policy for determining wing kinematics of a fly-mimicking flyer.
    In order to overcome the high computational cost of CFD-CSD that makes the coupling with deep-RL challenging, a novel data reproduction method is devised and the superiority of the method is demonstrated by comparing learning procedures with and without the reproduction method. 
    Furthermore, to learn the control policy valid in complex flow, multi-agent learning is utilized.
    In the learning, four flyers aviate in uniform flow with different speeds while sharing their data with a single deep-RL.
    Again, thanks to the data reproduction method, data of various directions can be obtained and, thus, the number of cases to learn is drastically reduced.
    After the learning, the learned policy is validated by showing its outstanding control performance in complex flow with abrupt disturbances.
    In addition, recovery behaviors of the flyer against the disturbances are investigated including the effect of the ambient flow field.
    
    In conclusion, the present study shows an excellent capability of a coupling framework of CFD-CSD and deep-RL as a means to generate a control policy for regulating highly intricate dynamics.
    Since the control policy is derived solely based on data from physical simulations, precise control is ensured.
    In addition, by virtue of CFD-CSD simulations, wing-wake interactions and behaviors of the flyer can be observed in detail, enhancing the understanding of the underlying physics and control mechanisms of the flyer.
    The next research topic could be an extension to three-dimensional flyer control to move a step forward to practical applications.
    In fact, the present framework can be easily extended to three dimensions as discussed in~\ref{sec:3D}, although further improvement in accelerating three-dimensional CFD-CSD is still required for immediate applications.
    In the same vein, the factors such as sensor delay or signal noise that can occur in real control processes should be considered.
    
    \section*{Declaration of competing interest}
	The authors declare that they have no known competing financial interests or personal relationships that could have appeared to influence the work reported in this paper.
    
    \section*{Acknowledgements}
	This work was supported by the National Research Foundation of Korea under Grant Number NRF-2021R1A2C2092146 and the Samsung Research Funding Center of Samsung Electronics under Project Number SRFC-TB1703-51.
	
	%% Figures %%
	\newpage
	%\listoffigures
	
	\pagebreak
	\clearpage
	
	\pagebreak
	\clearpage	
	
	\pagebreak
	\clearpage	

	\pagebreak
	\clearpage

    \pagebreak
    \clearpage

	\pagebreak
	\clearpage
    
    \pagebreak
	\clearpage
    \appendix
    
    \section{Methods} \label{sec:Methods}

    \subsection{Non-dimensionalization of variables} \label{sec:NV}
    In the present study, all variables are non-dimensionalized by reference scales. Dimensional and non-dimensional variables are differentiated depending on whether a mathematical font is used or not. The reference scales are determined based on the scales of a fruit fly~\cite{Shyy2010}. The reference length is the chord length of the wing $\mathpzc{c}=0.75mm$. The reference time is $1/2\mathpzc{f_r}$ where $\mathpzc{f_r}=200Hz$ is the reference flapping frequency of the wing. The reference velocity and the reference angular velocity are $\pi\mathpzc{f_r}\mathpzc{A_r}$ and $\pi\mathpzc{f_r}/3$, respectively, where $\mathpzc{A_r}=3.6\mathpzc{c}$ is the reference flapping amplitude of the wing. The reference force and the reference moment are $4\rho_\mathpzc{f}\mathpzc{c}^4\mathpzc{f_r}^2$ and $4\rho_\mathpzc{f}\mathpzc{c}^5\mathpzc{f_r}^2$, respectively, where $\rho_\mathpzc{f}=1.2\times10^{-6}g/mm^3$ is the density of the fluid. 
    
    \subsection{Simulation for solving the dynamics of a fly-mimicking flyer} \label{sec:FSI}
\begin{figure}
	\centering
	\includegraphics[width=\textwidth]{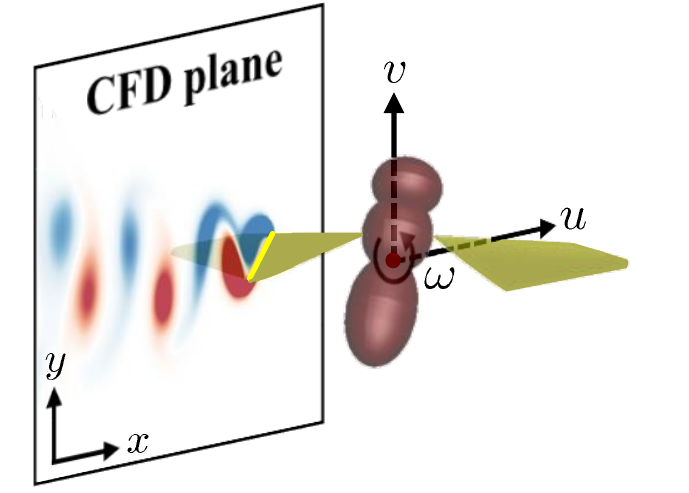}
	\caption{
    	Embodiment of a fly-mimicking flyer and its dynamics. The yellow line indicates a two-dimensional flexible wing in the domain where a CFD simulation is conducted. The red dot at the center of the flyer refers to the center of mass. The flyer moves parallel to the CFD plane with two translational and one rotational degrees of freedom.
	}
	\label{fig:CompConfig}
    \end{figure}

    Fig.~\ref{fig:CompConfig} shows the embodiment of a fly-mimicking flyer.
    To embody the flyer, a two-dimensional flexible wing is utilized and it flaps in a two-dimensional computational domain.
    The kinematics of the wing is described in~\ref{subsec:WK}.
    In the computational domain, a computational fluid dynamics (CFD) simulation is performed for fluid flow and the aerodynamic force exerted on the wing is calculated, which is further described in~\ref{subsec:NSE}. 
    Then, two assumptions are made to determine the total aerodynamic force on the flyer.
    The first assumption is that the aerodynamic force on the wing in the two-dimensional domain is the span-averaged value~\cite{Shyy2010}.
    The second assumption is that the aerodynamic force on the body of the flyer is neglected~\cite{Nguyen2016}.
    Thus, the total aerodynamic force on the flyer is calculated by multiplying the aspect ratio of the wing to the aerodynamic force on the two-dimensional wing.
    The aspect ratio of the wing is postulated as $A_R=\mathpzc{b}/\mathpzc{c}=6$ based on Shyy \textit{et al.}~\cite{Shyy2010} where $\mathpzc{b}$ is the span length of the wing.
    With the calculated total aerodynamic force, the dynamic equations of the flyer are solved by assuming that the body of the flyer has the shape of a cylinder with the length of $\mathpzc{L_b}=3\mathpzc{c}$, the radius of $\mathpzc{r_b}=\mathpzc{c}$, and the mass of $\mathpzc{m_b}=0.6mg$~\cite{Burggren2017}.
    The mass of the wing is postulated to be negligible compared to the mass of the body.
    The dynamic equations of the flyer are described in~\ref{subsec:BE}.
    In addition, to reflect the passive deformation of the wing, a computational structural dynamics (CSD) simulation is performed with the aerodynamic force on the wing, which is described in~\ref{subsec:WE} in detail. All the procedures are coupled to form an integrated computational fluid-structural dynamics (CFD-CSD) simulation, which is described in~\ref{subsec:Coupling}.
    
    \subsubsection{Kinematics of the wing} \label{subsec:WK}
\begin{figure}
	\centering
	\includegraphics[width=\textwidth]{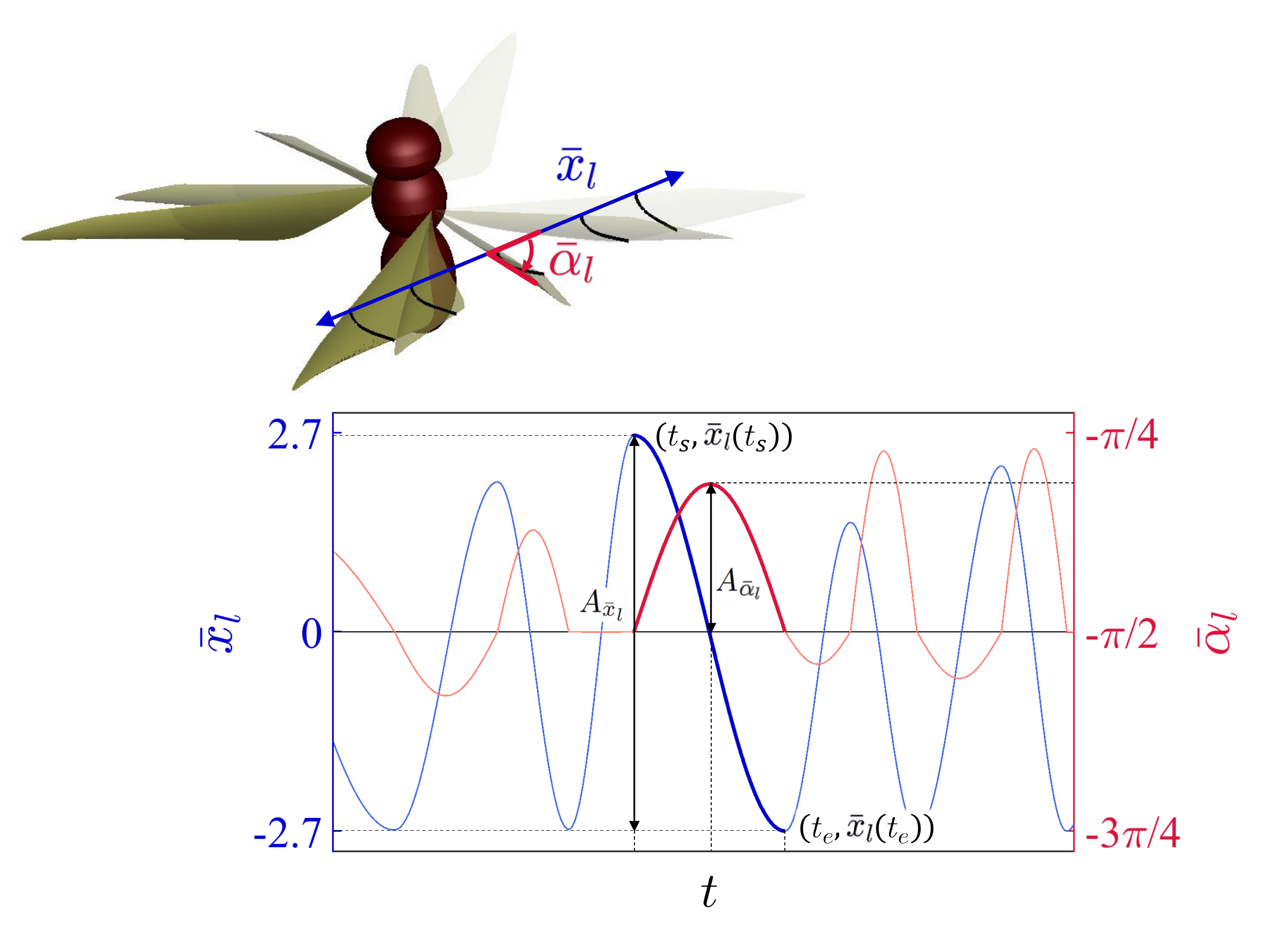}
	\caption{
    	Kinematics of a flyer's wing. At each stroke, the flyer determines the flapping amplitude $A_{\widebar{x}_l}$, the pitch amplitude $A_{\widebar{\alpha}_l}$, and the flapping frequency $f$ at the leading edge of the wing. As an example, the graph shows the time history of the wing kinematics where ${t}_s$ and ${t}_e$ refer to the times at the beginning and the end of a certain stroke, respectively.
	} 
	\label{fig:WingKine}
    \end{figure}

    Kinematics of the wing in the two-dimensional computational domain is described in this section.   
    Fig.~\ref{fig:WingKine} shows the schematic of the wing kinematics.
    The wing clamped at the leading edge moves under translational and rotational prescribed motions.
    The leading-edge motion at a certain stroke is described as follows:
    \begin{alignat}{2}
    \widebar{x}_l(t) &= \widebar{x}_l(t_s)-k(A_{\widebar{x}_l}/2)(cos(2\pi f(t-t_s))-1),
    \label{eqS3}\\
    \widebar{\alpha}_l(t) &= -\pi /2 - kA_{\widebar{\alpha}_l} sin(2\pi f(t-t_s)),
    \label{eqS4}
    \end{alignat}
    where $\widebar{x}_l(t)$ and $\widebar{\alpha}_l(t)$ represent the position and the pitch angle of the wing at the leading edge, respectively.
    The overbars indicate that the variables are represented with respect to the body-fixed frame.
    $k$ defines the binary state of a stroke where an upstroke corresponds to $k=1$ and a downstroke corresponds to $k=-1$. 
    $A_{\widebar{x}_l}$ refers the flapping amplitude which is the distance traveled by the leading edge during a stroke $|\widebar{x}_l(t_e)-\widebar{x}_l(t_s)|$, where $t_s$ and $t_e$ are the times at the beginning and the end of a stroke, respectively.
    Since the range of $\widebar{x}_l(t)$ is set to $[-2.7,2.7]$ according to Shyy \textit{et al}.~\cite{Shyy2010}, the range of $A_{\widebar{x}_l}$ is $[0,5.4]$.
    $A_{\widebar{\alpha}_l}$ refers the pitch amplitude of the leading edge, which has the range of $[0,\pi/4]$ in accordance with Yin and Luo~\cite{Yin2010}. 
    Lastly, $f$ is the flapping frequency normalized by the reference flapping frequency $\mathpzc{f_r}=200Hz$, ranging from $0.5$ to $1.5$. 
    These parameters completely define the position and angle of the wing at the leading edge.
    The position and angle at the rest part of the wing are obtained by solving the dynamic equation for the wing which is described in~\ref{subsec:WE}.
    
    \subsubsection{Navier-Stokes equations for fluid flow} \label{subsec:NSE}
    Flow fields in the computational domain are obtained by solving the non-dimensionalized incompressible Navier-Stokes equations given as follows: 	
    \begin{equation}
    \frac{\pd \bf{u}}{\pd t} + \bf{u}\cdot\nabla{\bf{u}} = -\nabla \text{$p$} + \frac{\text{$1$}}{\text{$Re$}}\nabla^2\bf{u}, 
    \label{eqS1}
    \end{equation}
    \begin{equation}
    {\nabla\cdot\bf{u}} - q = 0,
    \label{eqS2}
    \end{equation}
    where $\bf{u}$ and $p$ represent the velocity and the pressure of the fluid, respectively. The pressure is normalized by $\rho_\mathpzc{f}\mathpzc{U^2_r}$. The Reynolds number is defined as $Re=\mathpzc{U_r c}$/$\nu=84.8$ where $\mathpzc{U_r}=\pi\mathpzc{f_r}\mathpzc{A_r}$ is the reference velocity and $\nu$ is the kinematic viscosity of the fluid. $q$ in~\eqref{eqS2} is the mass source/sink term introduced to better conserve the mass of the fluid in the vicinity of the wing~\cite{KimKimChoi2001}.
    The Crank-Nicolson scheme is used for both convective and diffusion terms and a second-order central difference scheme is used for spatial discretization.
    In particular, the present study solves flow fields in the vicinity of the infinitesimally thin wing where special attentions should be paid. The detailed procedures of solving the above equations can be found in Hong \textit{et al}.~\cite{Hong2021}.
     
    Once a flow field is obtained, the total aerodynamic force and moment acting on the flyer are calculated. 
    The pressure force $\pmb{\mathpzc{F}}_\mathpzc{p}$ and the viscous force $\pmb{\mathpzc{F}}_\mathpzc{v}$ acting on an infinitesimal surface of the wing $d\mathpzc{S}$ can be represented as follows:
    \begin{alignat}{2}
    d\pmb{\mathpzc{F}}_\mathpzc{p} &= -{\rho_\mathpzc{f}}\mathpzc{p}\pmb{n}d\mathpzc{S},\\
    d\pmb{\mathpzc{F}}_\mathpzc{v} &= 2\mu\pmb{\mathpzc{D}}\cdot\pmb{n}d\mathpzc{S},
    \end{alignat}
    where $\mu$ is the dynamic viscosity of the fluid. $\pmb{\mathpzc D}$ is the strain-rate tensor defined as $(\nabla{\pmb{\mathpzc{u}}}+{\nabla{\pmb{\mathpzc{u}}}}^T)/2$ and $\pmb{n}$ is the outward normal vector of the surface. 
    The total aerodynamic force $\pmb{\mathpzc{F}}_a$ and moment ${\mathpzc{M}}_a$ acting on the flyer are computed as follows:
    \begin{alignat}{2}
    \pmb{\mathpzc{F}}_\mathpzc{a} &= \pmb{\mathpzc{F}}_\mathpzc{p} + \pmb{\mathpzc{F}}_\mathpzc{v} = \mathpzc{b}\int d\pmb{\mathpzc{F}}_\mathpzc{p} + \mathpzc{b}\int d\pmb{\mathpzc{F}}_\mathpzc{v},\\
    {\mathpzc{M}}_\mathpzc{a} &= {\mathpzc{M}}_\mathpzc{p} + {\mathpzc{M}}_\mathpzc{v} = \mathpzc{b}\int \pmb{r}\times d\pmb{\mathpzc{F}}_\mathpzc{p} + \mathpzc{b}\int \pmb{r}\times d\pmb{\mathpzc{F}}_\mathpzc{v},
    \end{alignat}
    where $\pmb{r}$ is the distance between the mass center of the flyer projected on the CFD plane and the infinitesimal surface of the wing.
    The obtained total aerodynamic force and moment on the flyer are used for solving the dynamics of the flyer in~\ref{subsec:BE}.
    
    \subsubsection{Dynamic equations for the flyer} \label{subsec:BE}
    The position and angle of the flyer are obtained by solving non-dimensionalized governing equations for the motion of the flyer as follows:
    \begin{equation}
    m_{bw}\frac{\pd^2{\bf X}_b}{\pd t^2} = \frac{{\bf F}_{a}}{A_Rh\rho_{wf}}+m_{bw}{\bf i}_{\bf g}Fr,
    \label{EOMB1}
    \end{equation}
    \begin{equation}
    I_{bw}\frac{\pd^2 {\theta}_b}{\pd t^2} = \frac{{M}_a}{A_Rh\rho_{wf}},
    \label{EOMB2}
    \end{equation}
    where $m_{bw}=\mathpzc{m_b}/\mathpzc{m_w}$ is the body to wing mass ratio.
    $\mathpzc{m_w}$ can be calculated as $\rho_w\mathpzc{bc}\mathpzc{h}$ where $\rho_w=1.2\times10^{-3}g/mm^3$ is the density of the wing and $\mathpzc{h}=7.5\times{10}^{-4}mm$ is the thickness of the wing~\cite{Kang2013}.
    ${{\bf{X}}_b}$ is the position of the flyer, ${\bf F}_a$ is the total aerodynamic force on the flyer, $\rho_{wf}=\rho_\mathpzc{w}/\rho_\mathpzc{f}$ is the wing to fluid density ratio, ${\bf i}_{\pmb{\mathpzc{g}}}={\pmb{\mathpzc{g}}}/\left|{\pmb{\mathpzc{g}}}\right|$ is the unit vector in the direction of gravity, and $Fr=\left|{\pmb{\mathpzc{g}}}\right|/4\mathpzc{cf^2_r}$ is the Froude number. In~\eqref{EOMB2}, $I_{bw}=\mathpzc{I}_\mathpzc{b}/\mathpzc{m_wc^2}$ is the moment of inertia of the body where $\mathpzc{I_b}=\mathpzc{m_b}\mathpzc{L^2_b}/12+\mathpzc{m_b}\mathpzc{r^2_b}/4$.
    ${\theta}_b$ is the pitch angle of the flyer, and ${M}_a$ is the total aerodynamic moment on the flyer.
    The equations are integrated in time using a second-order central difference scheme.
    
    \subsubsection{Dynamic equation for the wing} \label{subsec:WE}
    The passive deformation of the wing is obtained by solving the non-dimensionalized governing equation for the motion of the wing as follows:
    \begin{equation}
    \rho_{wf}\frac{\pd^2 {{\bf{X}}_w}}{\pd t^{2}} = \frac{\pd}{\pd s}\left(K_T\left(\left|\frac{\pd {\bf{X}}_w}{\pd s}\right|^2-1\right)\frac{\pd {\bf{X}}_w}{\pd s}\right)- K_B\frac{\pd^4 {{\bf{X}}_w}}{\pd s^4}+\frac{{\bf{{F}}}^*_a}{V}+\rho_{wf}{{\bf i}_{\bf g}}Fr, 
    \label{EOMW}
    \end{equation}
    where ${{\bf{X}}_w}$ is the position of the wing, $s$ is the Lagrangian coordinate along the two-dimensional wing, $K_T=\mathpzc{K_T}/4\rho_\mathpzc{f}\mathpzc{c^2}\mathpzc{hf^2_r}$ is the tension coefficient of the wing, $K_B=\mathpzc{K_B}/4\rho_\mathpzc{f}\mathpzc{c^4}\mathpzc{hf^2_r}$ is the bending coefficient of the wing, ${{\bf{F}}^*_a}$ is the aerodynamic force on the infinitesimal surface $\Delta s \cdot 1$ of the wing, and $V = \mathpzc{h}\Delta \mathpzc{s}\cdot1/\mathpzc{c^3}$ is the volume of each element with unit span length. 
    $K_T=7.4\times10^{6}$ and $K_B=2\times10^{4}$ are set similar to Kang and Shyy~\cite{Kang2013}.
    A second-order central difference scheme is used for both temporal integration and spatial discretization.
    
    \subsubsection{Coupling of computational fluid dynamics and computational structural dynamics} \label{subsec:Coupling}
    CFD and CSD are strongly coupled to ensure numerical stability and the coupling process is summarized as follows supposing that the time step advances from $n$ to $n+1$:
    \begin{enumerate}
    	\item At the time step $n+1$, for the first iteration $i=1$, ${\bf{X}}^{n+1,i}_b = {\bf{X}}^{n}_b$, ${\bf{X}}^{n+1,i}_w = {\bf{X}}^{n}_w$, and ${\bf{F}}^{n+1,i}_a={\bf{F}}^{n}_a$.	
    	\item The dynamic equations for the flyer (\eqref{EOMB1} and~\eqref{EOMB2}) and the wing (\eqref{EOMW}) are integrated in time to obtain ${\bf{X}}^{n+1,i+1}_b$, ${\theta}^{n+1,i+1}_b$, and ${\bf{X}}^{n+1,i+1}_w$.	
    	\item After moving the position of the flyer and the wing, ${\bf{X}}^{n+1,i+1}_w$, $\dot{\bf{X}}^{n+1,i+1}_w$, and $\ddot{\bf{X}}^{n+1,i+1}_w$ are provided as boundary conditions for CFD and ${\bf{u}}^{n+1,i+1}$ and $p^{n+1,i+1}$ are obtained.	
    	\item ${{\bf{F}}^{n+1,i+1}_a}$ is obtained from the updated flow field and the procedure is repeated over from step 2 until the convergence criteria ($|{\bf{X}}^{n+1,i+1}_w-{\bf{X}}^{n+1,i}_w|<$$10^{-8}$) is satisfied.
    \end{enumerate}
    While updating the position of the wing ${\bf{X}}^{n+1}_w$ from ${i}^{th}$ to ${(i+1)}^{th}$ iteration, the Aitken relaxation factor $\omega^{i}$~\cite{Kuttler2008} is used in order to ensure numerical stability of the staggered iteration as follows: 
    \begin{equation}
    {\bf{X}}^{n+1,i+1}_w = {\bf{X}}^{n+1,i}_w+\omega^{i}{\bf{r}}^{n+1,i}_w,
    \end{equation}
    where
    \begin{eqnarray*}
    	{\bf{r}}^{n+1,i}_w&=&{\tilde{\bf{X}}}^{n+1,i+1}_w-{\bf{X}}^{n+1,i}_w,\\	
    	{\omega}^{i}&=&-{\omega}^{i-1}\frac{({\bf{r}}^{n+1,i-1})^T({\bf{r}}^{n+1,i}-{\bf{r}}^{n+1,i-1})}{({\bf{r}}^{n+1,i}-{\bf{r}}^{n+1,i-1})^T({\bf{r}}^{n+1,i}-{\bf{r}}^{n+1,i-1})}. 
    \end{eqnarray*} 
    ${\tilde{\bf{X}}}^{n+1,i+1}_w$ is the predicted position of the wing by solving the dynamic equation (\eqref{EOMW}) for the wing at step 2.
    
    \subsection{Deep reinforcement learning} \label{sec:DRL}
    
    \subsubsection{Concept}
    Reinforcement learning is a process of learning a policy to determine an optimal action for an agent interacting with its environment~\cite{Sutton2018}.
    At each discrete step {\it n}, with a given state $s_n$, the agent selects an action $a_n$ according to its current policy; $a_n$ = $\pi(s_n)$.
    Then, the agent executes the action and, after interacting with the environment, gets a reward $r_n$ and the next state $s_{n+1}$.
    The objective of learning is to find the optimal policy which maximizes the value function $Q^{\pi}(s_n,a_n)$~\cite{Bellman1966}, defined as the expected sum of discounted future rewards as follows:
    \begin{equation}
    Q^{\pi}(s_n,a_n)={\mathop{\mathbb{E}}}(r_n+{\gamma}r_{n+1}+{\gamma}^2r_{n+2}+\dots), 
    \label{eq_valuefunctionq}
    \end{equation}
    where $\gamma$ is a discount factor that determines the weight between short-term and long-term rewards.
    
    If reinforcement learning is combined with deep learning, it is called deep reinforcement learning (Deep-RL).
    By incorporating deep neural networks in learning a control policy, deep-RL can handle complex and high-dimensional problems~\cite{mnih2015human}. 
    Deep-RL also has been successfully applied to various control problems by virtue of its generality~\cite{Bucsoniu2018}.
    Since deep-RL learns a control policy through the correlations between states, actions, and rewards, it does not require any equations to represent the dynamics of an agent. 
    Therefore, deep-RL can be easily applied to control problems that are too complex to formulate the dynamics of an agent.
    In the case of the present study, the dynamics of a fly-mimicking flyer is highly nonlinear due to the convoluted interaction with the ambient fluid.
    Therefore, as a method to derive an optimal control policy for a flyer, deep-RL is used.
    
    \subsubsection{State, action and reward}
    In the present section, the state, action, and reward of deep-RL are described.
    At the start of every wing stroke, the flyer receives a state and selects an action to be taken during the stroke. 
    The state at the $n^{th}$ stroke is defined as follows:
    \begin{equation}
    s_n = [{\bf{X}}_r,{\bf{\Theta}}_b,{\bf{U}},{\omega},{\bf{U}}_{a},\bar{{x}}_{l},k,{\bf{F}}_{b}],
    \label{EQ_State}
    \end{equation}
    where ${\bf{X}}_r=(x_r,y_r)=(x_b-x_{goal},y_b-y_{goal})$ and ${\bf{\Theta}}_b=(sin\theta_b,cos\theta_b)$ refer to the flyer's position with respect to the goal and the pitch angle of the flyer, respectively. 
    $sin\theta_b$ and $cos\theta_b$ are used to ensure the continuity of data~\cite{Brockman2016}.
    ${\bf{U}}=(u,v)$ and $\omega$ represent the velocity and the angular velocity of the flyer, respectively. 
    In addition, in order to control the flyer in the presence of external flow, the flyer has to sense the flow velocity. 
    Many insects including a fruit fly are known to perceive the external flow velocity through the antennae on their head~\cite{Fuller2014}. 
    Likewise, in the present study, an antenna is assumed to be positioned on the head of the flyer and it measures the flow velocity. 
    ${\bf{U}}_{a}=(u_a,v_a)$ is the probed flow velocity at the head of the flyer where $a$ in the subscript refers to the antenna. 
    $\widebar{x}_l(t)$ and $k$ represent the position of the wing at the leading edge and the binary state of the stroke, respectively.
    Lastly, ${\bf{F}}_b=(F_{bx},F_{by})$ refers to the body force (gravity) exerted on the flyer.
    The reason for including ${\bf{F}}_b$ is explained in detail in the main text. 
    
    The action determines the kinematics of the leading edge of the wing during the $n^{th}$ stroke, defined as follows:
    \begin{equation}
    a_n=[A_{\widebar{x}_l},A_{\widebar{\alpha}_l},f],
    \label{eq_action}
    \end{equation}
    where $A_{\widebar{x}_l}$, $A_{\widebar{\alpha}_l}$ and $f$ are the flapping amplitude, the pitch amplitude, and the flapping frequency, respectively.
    
    Lastly, the reward is the objective of control, defined as follows: 
    \begin{equation}
    r_n=-\sqrt{L/L_0}-{\omega}^2,
    \label{eq_reward}
    \end{equation}
    where $L=\sqrt{(x_b-x_{goal})^2+(y_b-y_{goal})^2}$ is the distance from the goal and $L_0$ is used for scaling, which is set to $1000$ in the present study.
    The aim of control is for the flyer to stably navigate to the specified goal and hover at that point.
    The use of $L$ makes the flyer to navigate to the goal and the square root is used to enhance the sensitivity of the distance near the goal.
    $\omega$ takes charge of stable flight of the flyer and the square is used to allow low angular velocities for changing its direction and simultaneously penalize unstable high angular velocities.
    All the variables included in the state, action, and reward are normalized by absolute magnitude around $1$ for better performance of learning. 
    
    \subsection{Data reproduction method} \label{sec:DRM}
    The basic concept and the advantage of the data reproduction method are introduced in the main text. 
    In this section, details on how the method is implemented in the present study are described.
    The original data to be reproduced have the form of $(s_n,a_n,r_n,s_{n+1})$.  
    There are three types of data reproduction methods: translation, mirroring, and rotation.
    For the translation, the reproduced $x_{start}$ and $y_{start}$ range from $-900$ to $900$ with the interval of $200$ where the computational domain size is $[-1000,1000]\times[-1000,1000]$. 
    Therefore, the number of the reproduced starting points is $100$. 
    At each reproduced starting point, the mirroring and rotation reproduction methods are conducted.
    Since the mirroring reproduction can be performed only in the $x$-direction, the number of data becomes twice. 
    For the rotation reproduction, the flyer is rotated with the interval of $\pi/5$, generating $10$ times more data.
    As a result, through the data reproduction method, the number of data become $2000$ times larger than that without the method.
    
    Table~\ref{tab:S1} shows how the components of a state change by the data reproduction method.
    $x_t$ and $y_t$ refer the translation distances and $\theta_r$ is the rotation angle.
    %In the table, for the mirroring reproduction, the signs of $\widebar{x}_l$ and $k$ change simultaneously.
    %This is because, in the two-dimensional domain, the flyer cannot turn back and forth.
    %Thus, in order to generate the data equivalent to mirroring-reproduced data within the domain, the body of the flyer is \red{assumed to be} flipped after mirroring, resulting in the change of the signs of $\widebar{x}_l$ and $k$. 
    In the case of an action, its components do not change because the kinematics of the wing does not change by the reproduction method.
    A reward is re-evaluated based on the reproduced state.
    
    %\raggedbottom
    \subsection{Learning procedure}
    \label{sec:LP}
    The learning procedure of the present study is summarized in Algorithm~\ref{alg_TD3}.
    For the deep-RL algorithm, the Twin Delayed Deep Deterministic policy gradient algorithm (TD3)~\cite{Fujimoto2018} is used, which is one of the state-of-the-art deep-RL algorithms.
    In TD3, two types of networks are introduced; actor and critic. 
    The actor network is a policy itself which determines an action in continuous space for a given state.
    The critic network predicts the value function.
    %Especially, by using two critic networks, TD3 greatly reduces the overestimation of the value function which has been often reported in the previous deep-RL algorithms.
    %In addition, TD3 considerably improves both the learning performance and stability through the two strategies: delayed policy update and target policy regularization, which can be seen in Algorithm~\ref{alg_TD3}.
    
    All of the neural networks are fully connected with four hidden layers of $400$, $400$, $400$, and $300$ neurons.
    The learning rates for the critic networks and the actor network are equally set to $10^{-4}$ and the Adam optimizer~\cite{Kingma2017} is used for updating the network parameters.
    The actor network and target networks are updated every 2 iterations ($d=2$) and the target update rate is set to $5\times10^{-4}$.
    The mini-batch size $N$ is set to $100$ and the discount factor $\gamma$ is set to $0.99$.
    The standard deviation of the exploration noise $\sigma$ is set to $1$ in the initial $100$ warm-up steps and $0.1$ afterward.
    If the flyer gets closer within the distance of $10$ from the goal, the noise is eliminated. 
    The standard deviation of the target regularization noise $\tilde{\sigma}$ is set to $0.2$ and the clip criteria $c_0$ is set to $0.5$.
    
    \subsection{Computational configurations} \label{sec:CC}
    Two computational domains are used in the present study. Except for the controlled flights in complex flow, the domain of $[-1000,1000]\times[-1000,1000]$ is used. 
    The computational grids are clustered near the wing and the total number of grids is $500\times500$. 
    The Neumann boundary condition is used at all the boundaries except for the multi-agent learning where uniform flow in the positive $x$-direction is imposed at the left boundary.
    
    On the other hand, while demonstrating the controlled flights in complex flow, the domain of $[-500,500]\times[-500,500]$ is set with the same grid numbers. 
    For complex flow, pulsatile flow is imposed at the left, bottom, and top boundaries. 
    The spatiotemporal variation of the flow is defined as follows:
    \begin{alignat}{3}
    u_{l} &= u_{lm}[1-0.5sin(2{\pi}f_l(t-t_i))(cos(2{\pi}(y+500)/L_y)-1)],\\
    v_{b} &= v_{bm}[1-0.5sin(2{\pi}f_b(t-t_i))(cos(2{\pi}(x+500)/L_x)-1)],\\   
    v_{t} &= -v_{b},
    \end{alignat} 
    where $l$, $b$, and $t$ represent left, bottom, and top, respectively. 
    $u_{lm}$ is set to $0.59$ and $v_{bm}$ is set to $0.29$, where the subscript $m$ refers to maximum amplitude. $f_l=0.01$ and $f_b=0.015$ refer to the frequencies of oscillations. $L_x=1000$ and $L_y=1000$ are the lengths of the domain.
    The flow speed at each boundary linearly increases from $0$ to its maximum amplitude within $t_i=40$ and starts to fluctuate as defined in the above equations. 
    The subscript $i$ refers to the initial time that the flow starts to fluctuate. 
    At the right boundary, the Neumann boundary condition is used.
    
    %% Figures %% 
    \pagebreak
	\clearpage

    \pagebreak
	\clearpage

    %% Table %%
	\pagebreak
	\clearpage
	\begin{table}\centering
	\caption{Changes in the components of a state by data reproduction methods}
	\begin{tabular}{p{0.17\textwidth}p{0.17\textwidth}p{0.17\textwidth}p{0.33\textwidth}}
		\hline
		Variables             & Translation  				 & Mirroring 				    & Rotation  \\
		\hline     
		$x_r$                   & $x_r+x_t$                     & $2x_{start}-x_r$          & $x_{start}+(x_r-x_{start})cos\theta_r-(y_r-y_{start})sin\theta_r$       \\
		$y_r$                   & $y_r+y_t$                     &                           & $y_{start}+(x_r-x_{start})sin\theta_r+(y_r-y_{start})cos\theta_r$       \\
		$sin\theta_b$         &                             & $sin(-\theta_b)$            & $sin(\theta_b+\theta_r)$ \\
		$cos\theta_b$         &                             & $cos(-\theta_b)$                             & $cos(\theta_b+\theta_r)$ \\
		${u}$                 &                             & $-u$                         & $ucos\theta_r-vsin\theta_r$ \\
		${v}$                 &                             &                              & $usin\theta_r+vcos\theta_r$ \\
		${\omega}$            &                             & $-\omega$                    &           \\
		${u}_{a}$             &                             & $-u_{a}$                     & $u_{a}cos\theta_r-v_{a}sin\theta_r$ \\
		${v}_{a}$             &                             &                              & $u_{a}sin\theta_r+v_{a}cos\theta_r$ \\	        
		$\widebar{x}_l$                 &                             & $-\widebar{x}_l$                       &           \\     
		$k$                   &                             & $-k$                         &           \\	      
		${F}_{bx}$            &                             & $-F_{bx}$                    & $F_{bx}cos\theta_r-F_{by}sin\theta_r$ \\    
		${F}_{by}$            &                             &                              & $F_{bx}sin\theta_r+F_{by}cos\theta_r$ \\
		\hline
	\end{tabular}\label{tab:S1}
    \end{table}
    
    %% Algorithm %%
    \pagebreak
	\clearpage
    \begin{algorithm}[]
    \footnotesize
	\caption{TD3 for learning a control policy of a flyer}
	\label{alg_TD3}
	\begin{algorithmic}
		\State initialize critic networks $Q_{\theta_1}$, $Q_{\theta_2}$, and actor network $\pi_{\phi}$ with random parameters $\theta_1$, $\theta_2$, $\phi$;
		\State initialize target networks $\theta_1'\gets\theta_1$, $\theta_2'\gets\theta_2$, $\phi'\gets\phi$;
		\State initialize replay buffer $\mathcal{B}$;
		\State $n\gets 1;$
		\Repeat			
		\State receive state $s_n$ from CFD-CSD simulation;
		\State select action with exploration noise: $a_n\gets$clip$(\pi_{\phi}(s_n)+\epsilon,-1,1)$, $\epsilon\sim\mathcal{N}(0,\sigma^{2})$;
		\State send the action to CFD-CSD simulation;
		\State $m\gets0$;				
		\Repeat
		\State sample mini-batch of $N$ transitions $(s_i,a_i,r_i,s_{i+1})$ from $\mathcal{B}$;
		\State $\tilde{a}\gets$ clip$(\pi_{\phi'}(s_{i+1})+\epsilon,-1,1)$, $\epsilon\sim$ clip$(\mathcal{N}(0,\tilde{\sigma}^{2}),-c_0,c_0)$;
		\State $y\gets r_i+\gamma$min$_{j=1,2}Q_{\theta_j'}(s_{i+1},\tilde{a})$;
		\State update $\theta_j$ with the loss function $L_{\theta_j}\gets N^{-1}\sum(y-Q_{\theta_j}(s_i,a_i))^2$, $j=1,2$;				
		\If{$m$ mod $d$}
		\State update $\phi$ by the deterministic policy gradient
		\State \ \ \ \ $N^{-1}\sum\nabla_{a_i} Q_{\theta_1}(s_i,a_i)|_{a_i=\pi_{\phi}(s_i)}\nabla_{\phi}\pi_{\phi}(s_i)$;
		\State update target networks:
		\State \ \ \ \ $\theta_1'\gets\tau\theta_1+(1-\tau)\theta_1'$, $\theta_2'\gets\tau\theta_2+(1-\tau)\theta_2'$, $\phi'\gets\tau\phi+(1-\tau)\phi'$;				
		\EndIf
		\State $m\gets m+1;$	
		\Until{next state $s_{n+1}$ and reward $r_n$ are provided;}
		\State reproduce data based on real data $(s_n,a_n,r_n,s_{n+1})$;
		\State store the reproduced data in $\mathcal{B}$;
		\State $n\gets n+1;$					
		\Until{hovering succeeds;}
	\end{algorithmic}	
    \end{algorithm}
    
    \pagebreak
	\clearpage
    \section{Supplementary results} \label{sec:SR}
    \subsection{Supporting results of multi-agent learning} \label{sec:SRML}
    This section includes two supporting results regarding the multi-agent learning in the main text. 
    The first result shows unsuccessful learning using the antenna-sensed velocity.
    The second result demonstrates controlled flights in untrained flow conditions.
    
    \subsubsection{Learning with the antenna-sensed velocity} \label{sec:SRML1}
\begin{figure}
	\centering
	\includegraphics[width=\textwidth]{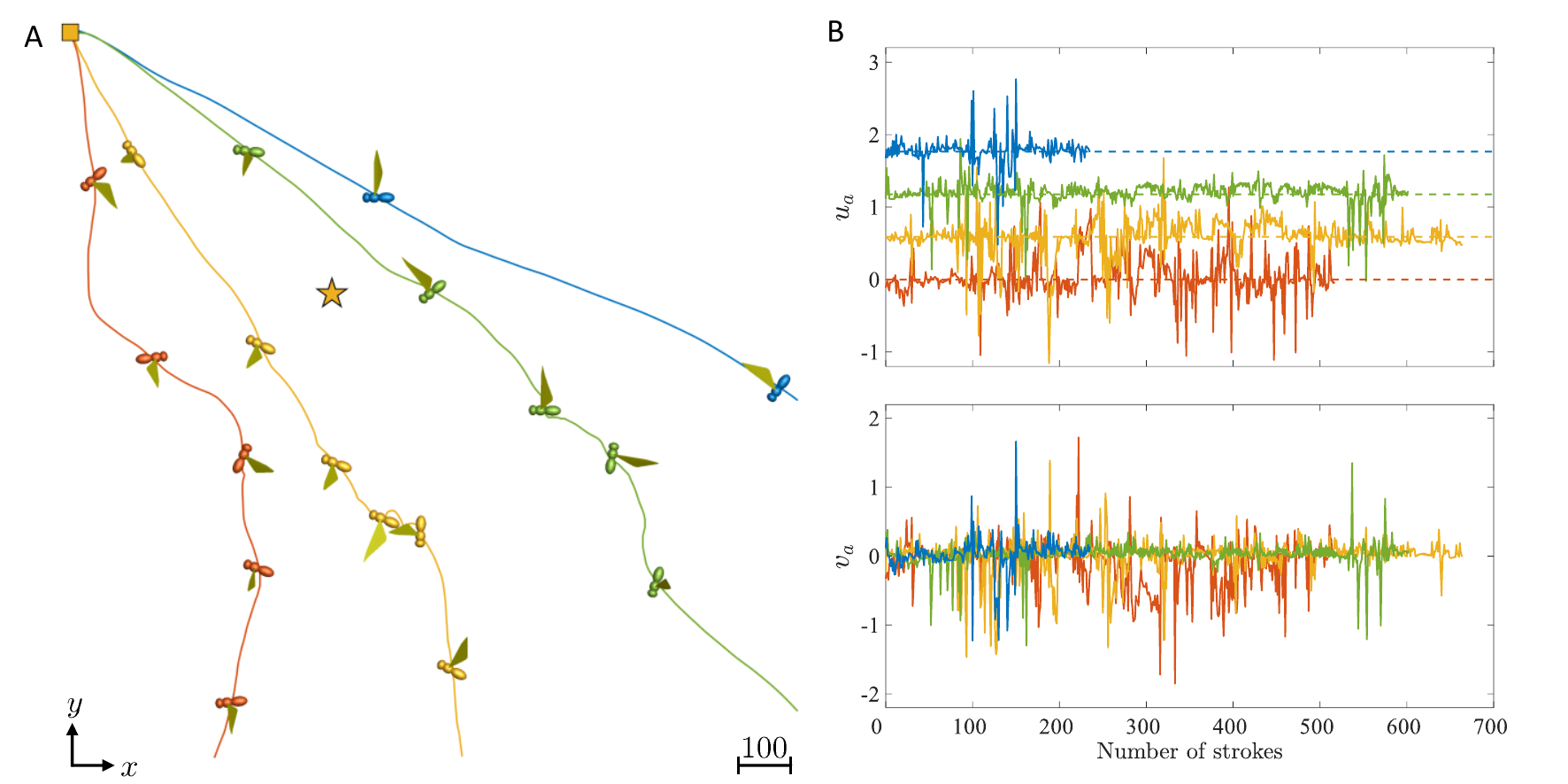}
	\caption{
    	Multi-agent learning with the antenna-sensed velocity in the positive $x$-directional uniform flow with four different speeds: \org{\linesolid}, 0; \yel{\linesolid}, 0.59; \ogr{\linesolid}, 1.18; \obl{\linesolid}, 1.76. (A) Trajectories of flyers. The flyers are depicted in 20 times magnification with the time interval of 100. (B) Antenna-sensed velocities of flyers. The solid and the dashed lines refer to the antenna-sensed velocities and the uniform flow velocities, respectively.
	}
	\label{fig:00ATN}
    \end{figure}

    Fig.~\ref{fig:00ATN} shows the result of the multi-agent learning where the only difference with Fig.~4 in the main text is using the antenna-sensed velocity as the state instead of the uniform flow velocity. 
    All the flyers fail to stabilize and go out of the domain helplessly.
    This is because the antenna-sensed velocity for each flyer shows very irregular patterns although the actual flow velocity is constant.
    This makes it challenging for the flyers to learn the environments.
    
    \subsubsection{Control in untrained flow conditions}\label{sec:SRML2}
\begin{figure}
	\centering
	\includegraphics[width=\textwidth]{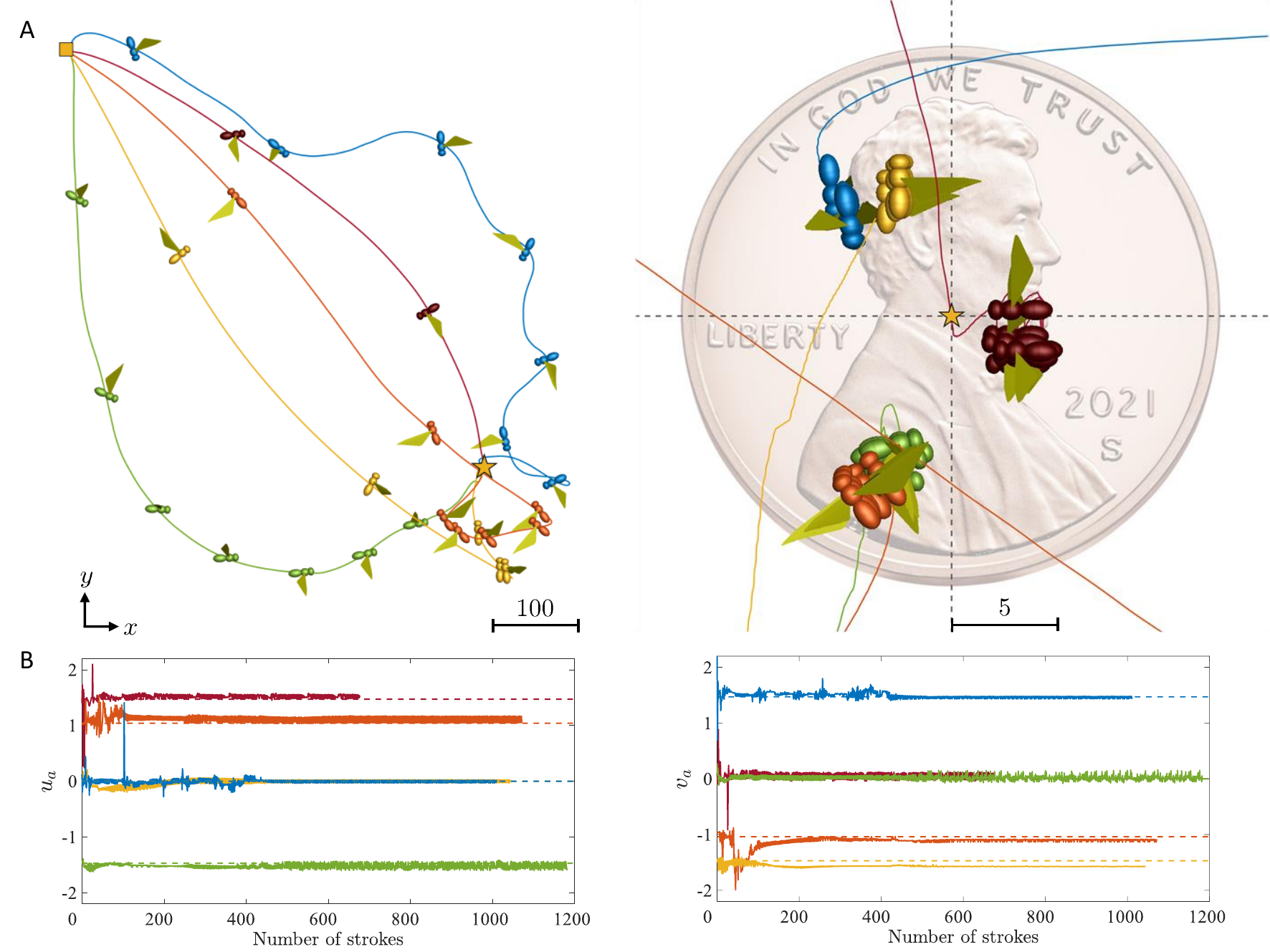}
	\caption{
    	Controlled flights of the flyers in untrained flow conditions. The flow speeds are the same as $1.47$ and various flow directions are tested:~\ord{\linesolid}, $(+x)$;~\org{\linesolid}, $(+x,-y)$ ($-45^{\circ}$ direction);~\yel{\linesolid}, $(-y)$;~\ogr{\linesolid}, $(-x)$;~\obl{\linesolid}, $(+y)$. (A) Trajectories of the flyers. For the overall view on the left, the flyers are depicted in 10 times magnification. Near the coin, the flyers are depicted in the real scale. All the flyers are depicted with the time interval of 50. (B) Antenna-sensed velocities of the flyers. The solid and the dashed lines refer to the antenna-sensed velocities and the uniform flow velocities, respectively.
	}
	\label{fig:25TEST}
    \end{figure}

    Fig.~\ref{fig:25TEST} concurrently supports the two arguments mentioned in the main text.
    One is if the learned flyer can deal with untrained flow conditions with different speeds and directions.
    The other is whether the difference in using the boundary-imposed velocity during learning and the antenna-sensed velocity during control arises any problem in control. 
    For testing the control performance, the five learned flyers are released in five untrained flow conditions.
    During the test, the antenna-sensed velocity is used for control and there is no additional learning.
    As shown in the figure, the flyers successfully aviate to the goal and hover for all the flow conditions.
    From the results, we can conclude that the learned policy can be applied to untrained flow conditions.
    In addition, after learning, since the flyers become dynamically stable, the antenna-sensed velocities give clear information to the policy. Therefore, the use of the boundary-imposed velocity while learning can be an adequate strategy.
    
    \subsection{Comparison on flyer's behaviors with and without external flow}\label{sec:CB}
\begin{figure}
	\centering
	\includegraphics[width=\textwidth]{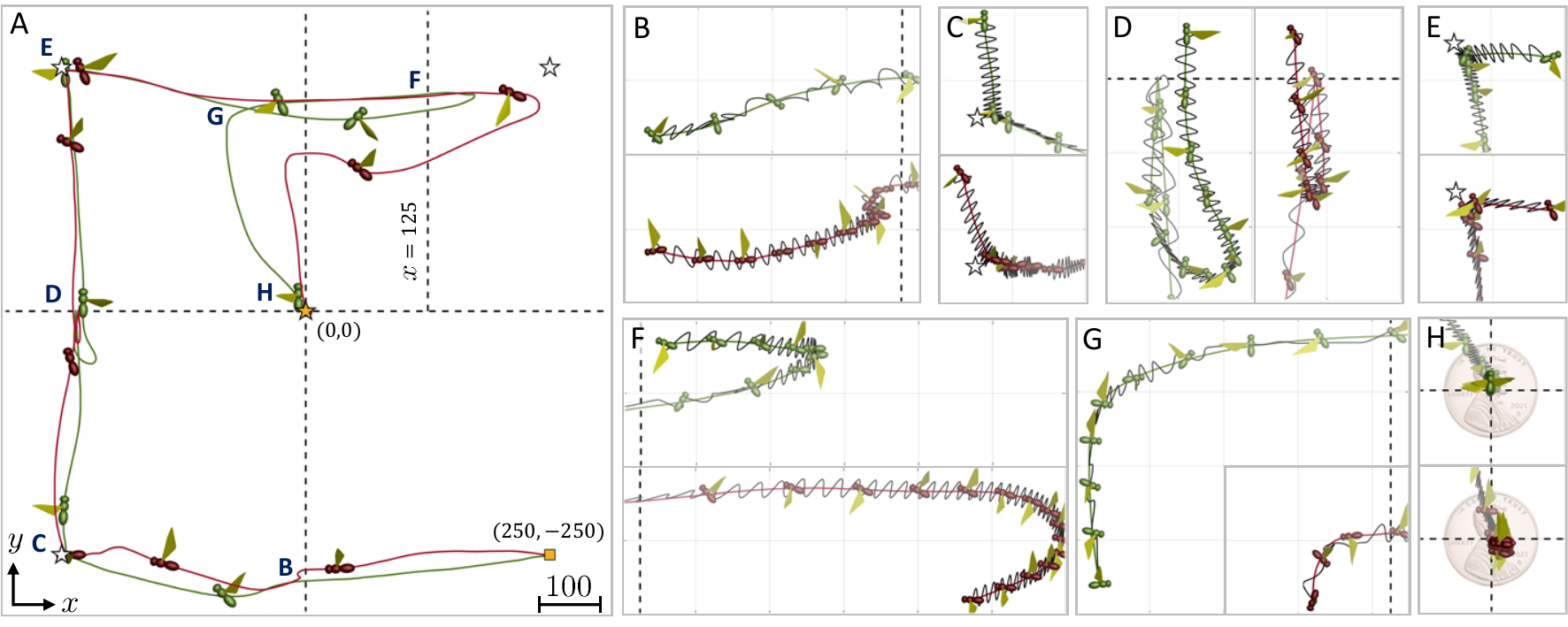}
	\caption{
	Comparison of the controlled flights of the flyers in complex flow (red) and a quiescent fluid (green). Trajectories of the flyers: (A) Overall trajectories of the flyers. (B) Recovery against the moment disturbance. (C) Direction change toward the updated goal. (D) Recovery against the force disturbance. (E) Direction change toward the updated goal. (F) $180^{\circ}$ turn. (G) $90^{\circ}$ turn. (H) Hovering. The flyers are depicted with different magnification factors and time intervals. For (A), the magnification factor is $10$ and the time interval is $100$. From (B) to (H), the magnification factor is $2$. For (B), (D), (F), and (G), the time interval is $5$. For (C), (E), and (H), the time interval is $15$. Note that, from (B) to (H), the grid spacing is $10$ and the trajectories become clearer as the flights proceed.
	}
	\label{fig:PulseFlyComp}
    \end{figure}

\begin{figure}[t]
	\centering
	\includegraphics[width=0.6\linewidth]{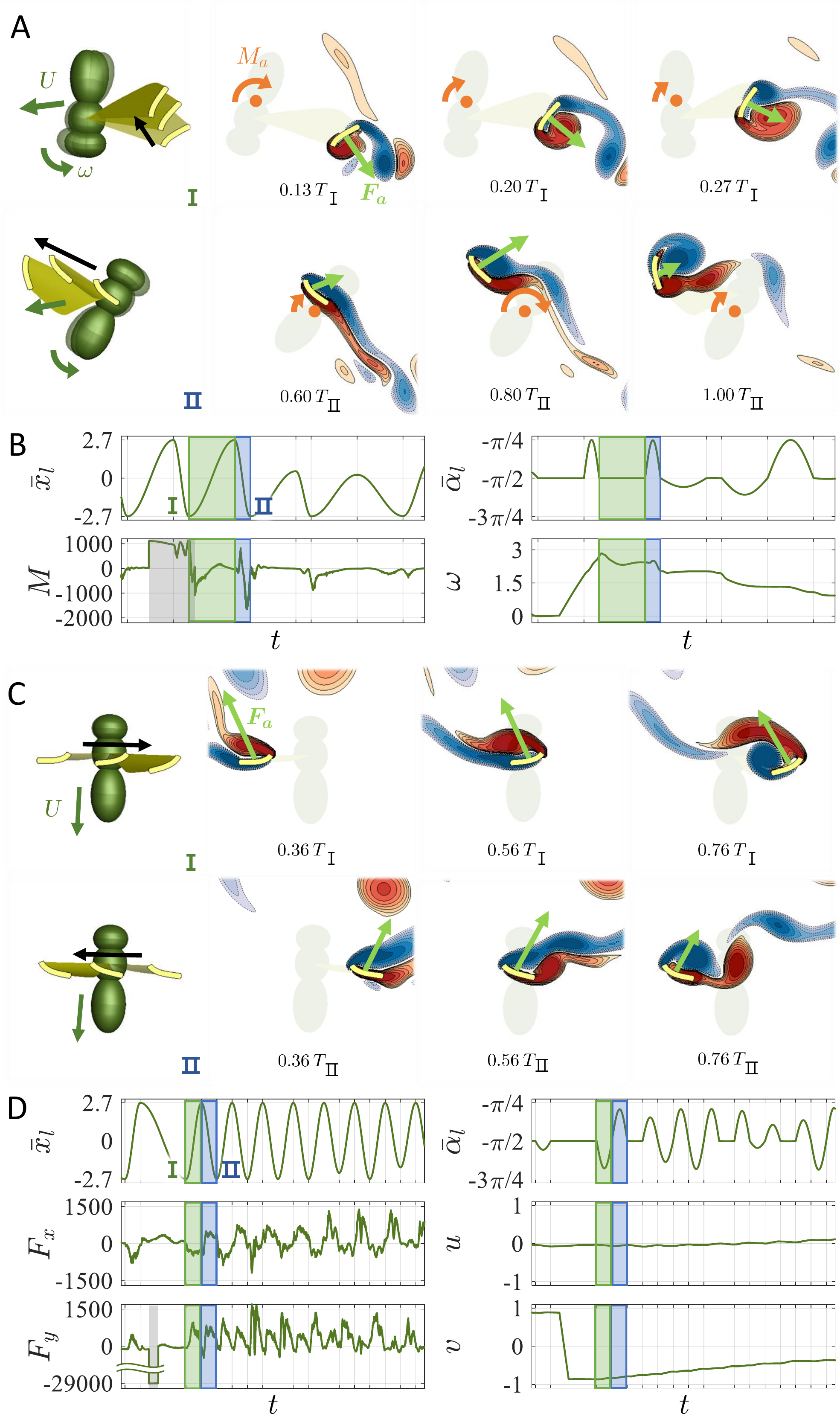}
	\caption{
	Analysis of recovery behaviors of the flyer in a quiescent fluid against disturbances. (A) Motions of the flyer and ambient flow fields corresponding to the part~\RNum{1} and~\RNum{2} in (B), respectively. (B) Time histories of the wing kinematics, total moment exerted on the flyer, and angular velocity of the flyer. (C) Motions of the flyer and ambient flow fields corresponding to the part~\RNum{1} and~\RNum{2} in (D), respectively. (D) Time histories of the wing kinematics, total $x$- and $y$-directional forces exerted on the flyer, and $x$- and $y$-directional velocities of the flyer. 
	The flow fields in (A) and (C) are depicted using vorticity where the red and blue colors refer to counter-clockwise and clockwise vortices, respectively.
	All the vectors are expressed in proportion to their magnitudes.
	$T_{\text{\RNum{1}}}$ and $T_{\text{\RNum{2}}}$ refer to the periods of time to take for the strokes~\RNum{1} and~\RNum{2}.
	The gray-colored parts in (B) and (D) refer to the durations of exerting the disturbances on the flyer: positive (counter-clockwise) body moment (B) and negative $y$-directional body force (D). $\bar{x}_l$ and $\bar{\alpha}_l$ represent the position and the pitch angle of the wing at the leading edge, respectively.
	} 
	\label{fig:TorqueRaindrop0ms}
    \end{figure}

    In the main text, a behavior analysis is conducted on the flyer controlled in complex flow.
    In order to exclude the effect of external flow and to investigate the behavioral differences, the same mission described in the main text is performed in a quiescent fluid and the discrepancies are dealt with in the present section.
    Fig.~\ref{fig:PulseFlyComp}A shows the overall flight trajectories of the flyers with and without external flow and both flyers successfully complete the mission.
    The figures from Fig.~\ref{fig:PulseFlyComp}B to Fig.~\ref{fig:PulseFlyComp}H represent the comparisons of the trajectories in the seven situations denoted in Fig.~\ref{fig:PulseFlyComp}A.
    There are some noticeable discrepancies when the flyers perform $180^{\circ}$ and $90^{\circ}$ turns which correspond to Figs.~\ref{fig:PulseFlyComp}F and~\ref{fig:PulseFlyComp}G, respectively.
    When the $180^{\circ}$ turn is performed, the flyer in external flow is distracted by the positive $x$-directional flow, resulting in the wider trajectory than the flyer in a quiescent fluid.
    On the other hand, when the $90^{\circ}$ turn is performed, the external flow assists the flyer to turn quickly toward the designated direction while the flyer in a quiescent fluid has to overcome its inertia by itself.
    %Thus, external flow may or may not be helpful depending on situations. 
    As can be seen in Figs.~\ref{fig:PulseFlyComp}C and~\ref{fig:PulseFlyComp}E, the flyer in external flow continuously heads toward the upstream direction, indicating that it is always conscious of the flow, while the flyer in a fluid quiescent fluid merely heads toward the goal. 
    Figs.~\ref{fig:PulseFlyComp}B and~\ref{fig:PulseFlyComp}D show the situations with disturbances, which is further described in detail in Fig.~\ref{fig:TorqueRaindrop0ms}.
    
    The recovery behaviors of the flyer in a quiescent fluid after the moment disturbance are shown in Figs.~\ref{fig:TorqueRaindrop0ms}A and~\ref{fig:TorqueRaindrop0ms}B.
    It shows similar behavioral patterns with the case of external flow described in the main text. 
    As shown in Fig.~\ref{fig:TorqueRaindrop0ms}A, although the velocity of the flyer is larger than that of the flyer in external flow, the relative velocity of the fluid experienced by the flyer is smaller due to the absence of external flow.
    This small relative velocity results in the smaller value of negative moment.
    Therefore, as one can compare the time histories of the angular velocities of the flyers in Fig.~\ref{fig:TorqueRaindrop0ms}B and Fig.~6B in the main text,
    the flyer in a quiescent fluid needs more strokes to stabilize its body whereas the flyer in external flow stabilizes only within two strokes.
    
    The recovery behaviors of the flyer in a quiescent fluid after the force disturbance are shown in Figs.~\ref{fig:TorqueRaindrop0ms}C and~\ref{fig:TorqueRaindrop0ms}D.
    As shown in Fig.~\ref{fig:TorqueRaindrop0ms}C, the motions of the wing for the upstroke and the downstroke are almost symmetric.
    This is different from what is observed for the flyer in external flow, where the upstroke and the downstroke show different behaviors taking separate roles.
    Since the flyer in a quiescent fluid only has to deal with the descending velocity, both strokes are used to recover the velocity generated from the disturbance.
    From the analysis, we can observe that the flyers appropriately behave depending on the confronted situations.
    
    \pagebreak
	\clearpage

    \pagebreak
	\clearpage

    \pagebreak
	\clearpage

    \pagebreak
	\clearpage
    
    \pagebreak
	\clearpage
    \section{Extension to a three-dimensional flyer}\label{sec:3D}
\begin{figure}
	\centering
	\includegraphics[width=\textwidth]{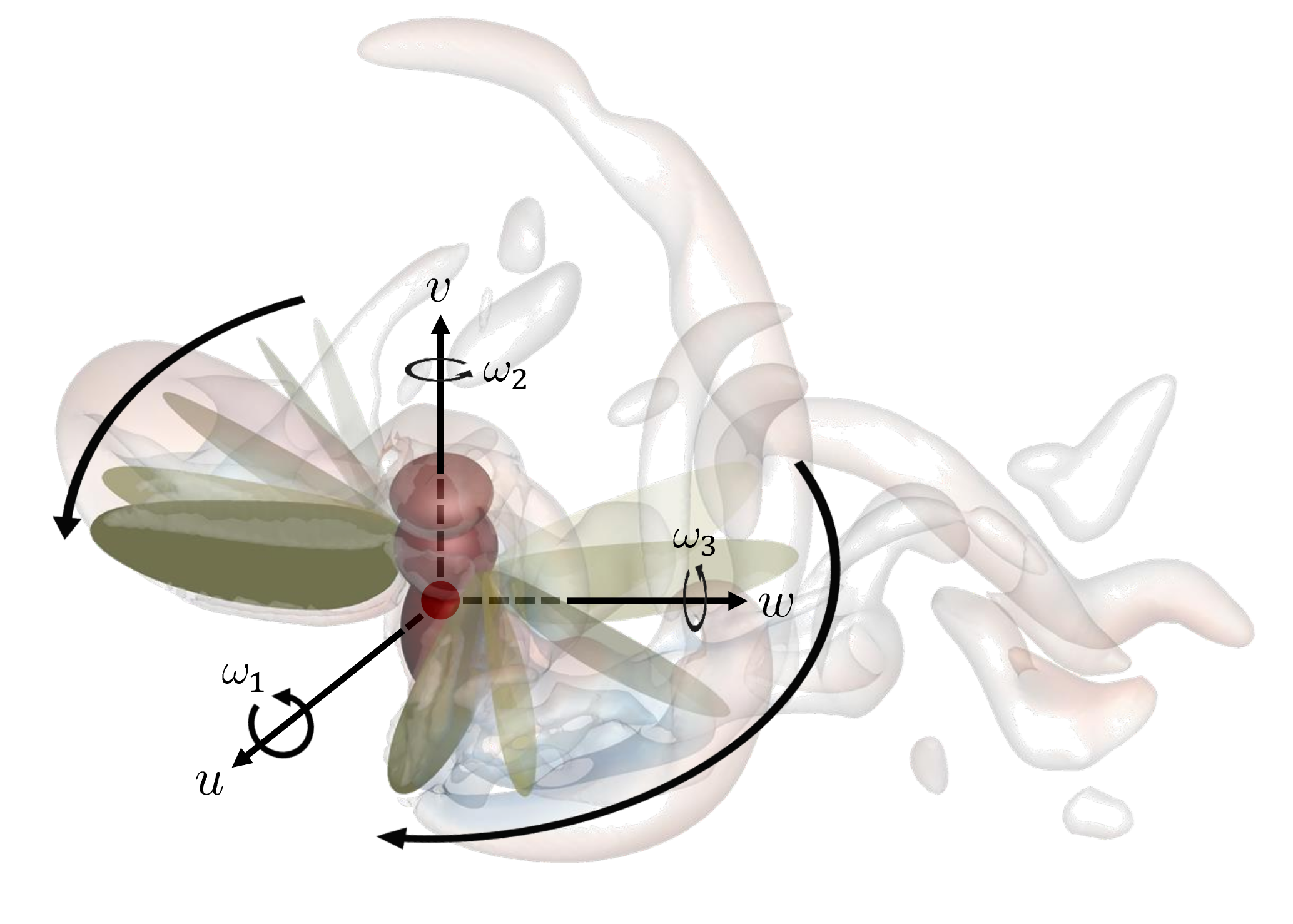}
	\caption{
    	Schematic of a three-dimensional flyer. The red dot denotes the center of mass of the flyer. Note that different active motions can be performed for each wing. As a demonstration, the flow field is calculated using the present CFD method. The vortical structure is depicted with $\lambda_2$-isosurface.
	}
	\label{fig:3DFlyer}
    \end{figure}

    The present study shows the superiority of an integrated framework of CFD-CSD and deep-RL for controlling a flyer in two dimensions. 
    In order to move a step forward to real applications, it should be extended to three dimensions. 
    In this section, a possible method for controlling a three-dimensional flyer is suggested, which is depicted in Fig.~\ref{fig:3DFlyer}.
    As shown in the figure, for a three-dimensional flyer, two wings have to be controlled individually.
    The state of deep-RL can be set as follows:   
    \begin{equation}
    s_n = [{\bf{X}}^{3D}_r,{\bf{\Theta}}^{3D}_b,{\bf{U}}^{3D},{\bf{\omega}}^{3D},{\bf{U}}^{3D}_{a},\bar{{\bf{X}}}^{3D}_l,k,{\bf{F}}^{3D}_{b}],
    \label{EQ_State_3D}
    \end{equation}
    where ${\bf{X}}^{3D}_r=(x_r,y_r,z_r)$ refers to the flyer's position relative to the goal. 
    ${\bf{\Theta}}^{3D}_b=(sin\theta_b,cos\theta_b,sin\phi_b,cos\phi_b,sin\psi_b,cos\psi_b)$ is the angle of the flyer where $\theta_b$, $\phi$, and $\psi$ refer to the pitch, roll, and yaw angles, respectively.
    ${\bf{U}}^{3D}=(u,v,w)$ and ${\bf{\omega}}^{3D}=(\omega_1,\omega_2,\omega_3)$ refer to the velocity and the angular velocity of the flyer, respectively. 
    ${\bf{U}}^{3D}_{a}=(u_a,v_a,w_a)$ is the probed flow velocity at the antenna of the flyer.
    $\bar{{\bf{X}}}^{3D}_l(t)=(\widebar{x}_{ll},\widebar{x}_{lr})$ represents the position of the wing at the leading edge where $l$ and $r$ in the second subscript refer to the left and the right wings, respectively.
    $k$ defines the binary state of a stroke.
    Note that only single $k$ is defined assuming the states of a stroke of both wings are the same for simplicity.
    ${\bf{F}}^{3D}_{b}=(F_{bx},F_{by},F_{bz})$ is the body force exerted on the flyer.
    An action determines the kinematics of the wings at the leading edges during a stroke, defined as follows:
    \begin{equation}
    a_n=[A_{\widebar{x}_{ll}},A_{\widebar{x}_{lr}},A_{\widebar{\alpha}_{ll}},A_{\widebar{\alpha}_{lr}},f],
    \label{EQ_action_3D}
    \end{equation}
    where $A_{\widebar{x}_{ll}}$, $A_{\widebar{x}_{lr}}$, $A_{\widebar{\alpha}_{ll}}$, $A_{\widebar{\alpha}_{lr}}$, and $f$ are the flapping amplitudes, pitch amplitudes for the left and right wings, and flapping frequency. 
    Note that only one frequency is used to synchronize the communication period between CFD-CSD and deep-RL.
    Lastly, a reward is defined as follows: 
    \begin{equation}
    r_n=-\sqrt{L/L_0}-|{\bf{\omega}}^{3D}|^2,
    \label{EQ_reward_3D}
    \end{equation}
    where $L=\sqrt{(x_b-x_{goal})^2+(y_b-y_{goal})^2+(z_b-z_{goal})^2}$ is the distance from the goal and
    $L_0$ is a scaling factor.   
    
    As shown in the present study, the data reproduction method can be used in the same manner in three dimensions.
    Although the number of variables and the corresponding relations to learn increase, more diverse data can be reproduced with the increased degree of freedom.
    
    \pagebreak
	\clearpage
    
	\pagebreak
    \clearpage
	%\section*{References}
    \bibliographystyle{elsarticle-num}
    \biboptions{sort&compress}
    \bibliography{FlyControl}

\begin{thebibliography}{10}
\expandafter\ifx\csname url\endcsname\relax
  \def\url#1{\texttt{#1}}\fi
\expandafter\ifx\csname urlprefix\endcsname\relax\def\urlprefix{URL }\fi
\expandafter\ifx\csname href\endcsname\relax
  \def\href#1#2{#2} \def\path#1{#1}\fi

\bibitem{Phan2019}
H.~V. Phan, H.~Park, Insect-inspired, tailless, hover-capable flapping-wing
  robots: Recent progress, challenges, and future directions, Progress in
  Aerospace Sciences 111 (2019) 100573.

\bibitem{Karasek2018Delfly}
M.~Kar{\'a}sek, F.~T. Muijres, C.~De~Wagter, B.~D. Remes, G.~C. de~Croon, A
  tailless aerial robotic flapper reveals that flies use torque coupling in
  rapid banked turns, Science 361 (2018) 1089--1094.

\bibitem{Fei2019LearnExtreme}
F.~Fei, Z.~Tu, J.~Zhang, X.~Deng, Learning extreme hummingbird maneuvers on
  flapping wing robots, in: 2019 International Conference on Robotics and
  Automation, 2019, pp. 109--115.

\bibitem{Phan2020Collision}
H.~V. Phan, H.~Park, Mechanisms of collision recovery in flying beetles and
  flapping-wing robots, Science 370~(6521) (2020) 1214--1219.

\bibitem{Cunis2016precision}
T.~Cunis, M.~Karasek, G.~de~Croon, Precision position control of the delfly
  {II} flapping-wing micro air vehicle in a wind-tunnel, in: The International
  Micro Air Vehicle Conference and Competition 2016, 2016, pp. 17--21.

\bibitem{Karasek2019accurate}
M.~Kar{\'a}sek, M.~Percin, T.~Cunis, B.~W. van Oudheusden, C.~De~Wagter, B.~D.
  Remes, G.~C. de~Croon, Accurate position control of a flapping-wing robot
  enabling free-flight flow visualisation in a wind tunnel, International
  Journal of Micro Air Vehicles 11 (2019) 1756829319833683.

\bibitem{Chirara2017}
P.~Chirarattananon, Y.~Chen, E.~F. Helbling, K.~Y. Ma, R.~Cheng, R.~J. Wood,
  Dynamics and flight control of a flapping-wing robotic insect in the presence
  of wind gusts, Interface Focus 7 (2017) 20160080.

\bibitem{Ma2013}
K.~Y. Ma, P.~Chirarattananon, S.~B. Fuller, R.~J. Wood, Controlled flight of a
  biologically inspired, insect-scale robot, Science 340~(6132) (2013)
  603--607.

\bibitem{Pulskamp2012}
J.~S. Pulskamp, R.~G. Polcawich, R.~Q. Rudy, S.~S. Bedair, R.~M. Proie,
  T.~Ivanov, G.~L. Smith, Piezoelectric {PZT} {MEMS} technologies for
  small-scale robotics and {RF} applications, MRS bulletin 37~(11) (2012)
  1062--1070.

\bibitem{Ellington1984}
C.~P. Ellington, The aerodynamics of hovering insect flight. {I}. the
  quasi-steady analysis, Philos. Trans. R. Soc. Lond., B, Biol. Sci. 305~(1122)
  (1984) 1--15.

\bibitem{Sane2003}
S.~P. Sane, The aerodynamics of insect flight, Journal of Experimental Biology
  206~(23) (2003) 4191--4208.

\bibitem{Bluman2017}
J.~Bluman, C.~Kang, Wing-wake interaction destabilizes hover equilibrium of a
  flapping insect-scale wing, Bioinspir. Biomim. 12~(4) (2017) 046004.

\bibitem{Ristroph2010}
L.~Ristroph, A.~J. Bergou, G.~Ristroph, K.~Coumes, G.~J. Berman,
  J.~Guckenheimer, Z.~J. Wang, I.~Cohen, Discovering the flight autostabilizer
  of fruit flies by inducing aerial stumbles, Proceedings of the National
  Academy of Sciences 107~(11) (2010) 4820--4824.

\bibitem{Ristroph2013}
L.~Ristroph, G.~Ristroph, S.~Morozova, A.~J. Bergou, S.~Chang, J.~Guckenheimer,
  Z.~J. Wang, I.~Cohen, Active and passive stabilization of body pitch in
  insect flight, Journal of the R. Society Interface 10~(85) (2013) 20130237.

\bibitem{Phan2017}
H.~V. Phan, T.~Kang, H.~Park, Design and stable flight of a 21 g insect-like
  tailless flapping wing micro air vehicle with angular rates feedback control,
  Bioinspir. Biomim. 12~(3) (2017) 036006.

\bibitem{Teoh2012}
Z.~E. Teoh, S.~B. Fuller, P.~Chirarattananon, N.~Prez-Arancibia, J.~D.
  Greenberg, R.~J. Wood, A hovering flapping-wing microrobot with altitude
  control and passive upright stability, in: 2012 IEEE/RSJ International
  Conference on Intelligent Robots and Systems, 2012, pp. 3209--3216.

\bibitem{Altshuler2005}
D.~L. Altshuler, W.~B. Dickson, J.~T. Vance, S.~P. Roberts, M.~H. Dickinson,
  Short-amplitude high-frequency wing strokes determine the aerodynamics of
  honeybee flight, Proceedings of the National Academy of Sciences 102~(50)
  (2005) 18213--18218.

\bibitem{Bomphrey2017}
R.~J. Bomphrey, T.~Nakata, N.~Phillips, S.~M. Walker, Smart wing rotation and
  trailing-edge vortices enable high frequency mosquito flight, Nature
  544~(7648) (2017) 92--95.

\bibitem{Kang2011}
C.-K. Kang, H.~Aono, C.~E. Cesnik, W.~Shyy, Effects of flexibility on the
  aerodynamic performance of flapping wings, Journal of Fluid Mechanics 689
  (2011) 32--74.

\bibitem{Bucsoniu2018}
L.~Bu{\c{s}}oniu, T.~de~Bruin, D.~Toli{\'c}, J.~Kober, I.~Palunko,
  Reinforcement learning for control: Performance, stability, and deep
  approximators, Annual Reviews in Control 46 (2018) 8--28.

\bibitem{Gazzola2014}
M.~Gazzola, B.~Hejazialhosseini, P.~Koumoutsakos, Reinforcement learning and
  wavelet adapted vortex methods for simulations of self-propelled swimmers,
  SIAM Journal on Scientific Computing 36~(3) (2014) B622--B639.

\bibitem{Verma2018}
S.~Verma, G.~Novati, P.~Koumoutsakos, Efficient collective swimming by
  harnessing vortices through deep reinforcement learning, Proceedings of the
  National Academy of Sciences 115~(23) (2018) 5849--5854.

\bibitem{Birch2001span1}
J.~M. Birch, M.~H. Dickinson, Spanwise flow and the attachment of the
  leading-edge vortex on insect wings, Nature 412 (2001) 729--733.

\bibitem{Shyy2007span2}
W.~Shyy, H.~Liu, Flapping wings and aerodynamic lift: the role of leading-edge
  vortices, AIAA Journal 45 (2007) 2817--2819.

\bibitem{Zhu2020Speed}
H.~J. Zhu, M.~Sun, Kinematics measurement and power requirements of fruitflies
  at various flight speeds, Energies 13 (2020) 4271.

\bibitem{Fuller2014}
S.~B. Fuller, A.~D. Straw, M.~Y. Peek, R.~M. Murray, M.~H. Dickinson, Flying
  drosophila stabilize their vision-based velocity controller by sensing wind
  with their antennae, Proceedings of the National Academy of Sciences 111~(13)
  (2014) E1182--E1191.

\bibitem{Dickerson2012}
A.~K. Dickerson, P.~G. Shankles, N.~M. Madhavan, D.~L. Hu, Mosquitoes survive
  raindrop collisions by virtue of their low mass, Proceedings of the National
  Academy of Sciences 109~(25) (2012) 9822--9827.

\bibitem{Shyy2010}
W.~Shyy, H.~Aono, S.~K. Chimakurthi, P.~Trizila, C.~Kang, C.~E. Cesnik, H.~Liu,
  Recent progress in flapping wing aerodynamics and aeroelasticity, Progress in
  Aerospace Sciences 46~(7) (2010) 284--327.

\bibitem{Nguyen2016}
A.~T. Nguyen, J.-K. Kim, J.-H. Han, The effects of body aerodynamics on the
  dynamic stability of insect flight, in: 30th Congress of the International
  Council of the Aeronautical Sciences, 2016.

\bibitem{Burggren2017}
W.~Burggren, B.~M. Souder, D.~H. Ho, Metabolic rate and hypoxia tolerance are
  affected by group interactions and sex in the fruit fly (drosophila
  melanogaster): new data and a literature survey, Biology Open 6 (2017)
  471--480.

\bibitem{Yin2010}
B.~Yin, H.~Luo, Effect of wing inertia on hovering performance of flexible
  flapping wings, Physics of Fluids 22~(11) (2010) 111902.

\bibitem{KimKimChoi2001}
J.~Kim, D.~Kim, H.~Choi, An immersed-boundary finite-volume method for
  simulations of flow in complex geometries, Journal of Computational Physics
  171 (2001) 132--150.

\bibitem{Hong2021}
S.~Hong, D.~Yoon, S.~Ha, D.~You, A ghost-cell immersed boundary method for
  unified simulations of flow over finite-and zero-thickness moving bodies at
  large cfl numbers, Engineering Applications of Computational Fluid Mechanics
  15~(1) (2021) 437--461.

\bibitem{Kang2013}
C.~Kang, W.~Shyy, Scaling law and enhancement of lift generation of an
  insect-size hovering flexible wing, Journal of the R. Society Interface 10
  (2013) 20130361.

\bibitem{Kuttler2008}
U.~K{\"u}ttler, W.~A. Wall, Fixed-point fluid--structure interaction solvers
  with dynamic relaxation, Computational Mechanics 43~(1) (2008) 61--72.

\bibitem{Sutton2018}
R.~S. Sutton, A.~G. Barto, Reinforcement {L}earning: An {I}ntroduction, MIT
  Press, Cambridge, MA, 1998.

\bibitem{Bellman1966}
R.~Bellman, Dynamic programming, Science 153~(3731) (1966) 34--37.

\bibitem{mnih2015human}
V.~Mnih, K.~Kavukcuoglu, D.~Silver, A.~A. Rusu, J.~Veness, M.~G. Bellemare,
  A.~Graves, M.~Riedmiller, A.~K. Fidjeland, G.~Ostrovski, et~al., Human-level
  control through deep reinforcement learning, nature 518~(7540) (2015)
  529--533.

\bibitem{Brockman2016}
G.~Brockman, V.~Cheung, L.~Pettersson, J.~Schneider, J.~Schulman, J.~Tang,
  W.~Zaremba, Open{AI} gym, ar{X}iv
  [Preprint]\url{https://arxiv.org/abs/1606.01540v1} (accessed 23 April 2021).

\bibitem{Fujimoto2018}
S.~Fujimoto, H.~Hoof, D.~Meger, Addressing function approximation error in
  actor-critic methods, in: International Conference on Machine Learning, 2018,
  pp. 1587--1596.

\bibitem{Kingma2017}
D.~P. Kingma, J.~Ba, Adam: A method for stochastic optimization, ar{X}iv
  [Preprint]\url{https://arxiv.org/abs/1412.6980v9} (accessed 7 March 2021).

\end{thebibliography}
    
\pagebreak
\clearpage

\end{document}